# Evidence Is All You Need: Ordering Imaging Studies via Language Model Alignment with the ACR Appropriateness Criteria

**Michael S. Yao**[1,2], **Allison Chae**[2], **Charles E. Kahn, Jr.**[2,3], **Walter R. Witschey**[2,3], **James C. Gee**[3], **Hersh Sagreiya**[2,3,†], **Osbert Bastani**[4,†,*]
[1]Department of Bioengineering, University of Pennsylvania, Philadelphia, PA 19104
[2]Perelman School of Medicine, University of Pennsylvania, Philadelphia, PA 19104
[3]Department of Radiology, University of Pennsylvania, Philadelphia, PA 19104
[4]Department of Computer and Information Science, University of Pennsylvania, Philadelphia, PA 19104
[†]Denotes equal contribution.
*Correspondence to OB (obastani@seas.upenn.edu)

**ABSTRACT**
Diagnostic imaging studies are an increasingly important component of the workup and management of acutely presenting patients. However, ordering appropriate imaging studies according to evidence-based medical guidelines is a challenging task with a high degree of variability between healthcare providers. To address this issue, recent work has investigated if generative AI and large language models can be leveraged to help clinicians order relevant imaging studies for patients. However, it is challenging to ensure that these tools are correctly aligned with medical guidelines, such as the American College of Radiology's Appropriateness Criteria (ACR AC). In this study, we introduce a framework to intelligently leverage language models by recommending imaging studies for patient cases that are aligned with evidence-based guidelines. We make available a novel dataset of patient "one-liner" scenarios to power our experiments, and optimize state-of-the-art language models to achieve an accuracy on par with clinicians in image ordering. Finally, we demonstrate that our language model-based pipeline can be used as intelligent assistants by clinicians to support image ordering workflows and improve the accuracy of imaging study ordering according to the ACR AC. Our work demonstrates and validates a strategy to leverage AI-based software to improve trustworthy clinical decision making in alignment with evidence-based guidelines.

**INTRODUCTION**
Inappropriate ordering of diagnostic imaging studies is a commonly encountered problem in the emergency department and other acute-care settings.[1–4] According to the American College of Radiology (ACR), up to 50% of imaging studies performed in the ED every year are not clinically indicated, often resulting in unnecessary radiation exposure for patients and significant administrative and financial burden for hospital systems.[5] To address this problem, the ACR published the ACR Appropriateness Criteria® (ACR AC) to provide evidence-based guidelines that assist referring physicians in ordering the most appropriate diagnostic imaging studies for specific clinical conditions.[6] As of its most recent online release, the ACR AC contains 263 unique imaging topics (i.e., patient scenarios).

However, despite the widespread availability of the ACR AC to clinicians, improper imaging according to the guidelines remains a challenge in many emergency departments and inpatient settings.[3,5] Bautista et al. showed that there is low utilization of the ACR AC by clinicians in practice: less than 1% of physicians interviewed in their study use the ACR AC as a first-line resource when ordering diagnostic imaging studies.[7,8] The limited usage of the ACR AC may be partly due to how the Appropriateness Criteria are presented to clinicians; the evidence-based criteria are dense and can be difficult to parse through even for physician experts, and especially in acute healthcare settings such as the emergency department where decision making is both time-sensitive and critical.

To address this problem, recent work has investigated the potential utility of generative artificial intelligence (AI) tools to synthesize dense passages of evidence-based guidelines to offer clinical decision support (CDS) in physician workflows.[9–11] In particular, large language models (LLMs) are generative AI models trained on large corpora of textual data to achieve impressive performance on tasks such as language translation, summarization, and text generation.[12–15] In principle, sophisticated AI tools such as LLMs could be used to recommend diagnostic imaging studies in alignment with the ACR AC evidence-based guidelines based on input summaries of a patient presentation. However, despite the success of these models in natural language tasks, LLMs have been shown to struggle in challenging *domain-specific* tasks requiring human expertise and specialized training, such as in medicine, law, and engineering.[16–18] As a result, accessing the potential benefits of LLMs in these domains—such as for recommending appropriate imaging studies for patients—continues to be an ongoing challenge, deterring widespread adaptation of generative AI models in clinical medicine.[19,20]

In this work, we investigate how state-of-the-art LLMs can be used as potential CDS tools to reduce the problem over-imaging according to the ACR AC. Our core hypothesis is that while LLMs may struggle to directly recommend imaging studies for patients (a domain-specific task), they are often able to accurately describe patient conditions and presentations. In this light, we apply LLMs to analyze patient "one-liner" summaries and map them to topic categories from the ACR AC. We can then programmatically query the ACR AC based on the LLM-recommended topic category (without any explicit LLM usage) to determine the optimal imaging study for a patient. In this fashion, LLMs can be used to recommend diagnostic imaging studies according to recommendations from the guidelines.

Our contributions are as follows: firstly, we introduce **RadCases**, the first publicly available dataset of one-liners labelled by the most relevant ACR AC panel and topic. Secondly, we evaluate publicly available, state-of-the-art LLMs on our RadCases dataset to characterize how existing tools may be used out-of-the-box for diagnostic imaging support in inpatient settings. We then assess how popular techniques such as model fine-tuning, retrieval-augmented generation (RAG), in-context learning (ICL), and chain-of-thought prompting (COT) may be effectively leveraged to improve the alignment of existing LLMs with ACR AC, and also enable LLMs to perform comparably to clinicians in image ordering in a retrospective analysis.[21,22] Finally, we conduct a prospective randomized controlled trial to demonstrate that LLM clinical assistants can improve image ordering by clinicians in acute care settings.

## RESULTS
### Constructing a benchmarking dataset for image ordering using language models.
Prior work in medical natural language processing have primarily focused on tasks such as documentation writing, medical question answering, and chatbot-clinician alignment. In each of these tasks, a relatively complete picture including hospital course, lab values, and advanced image studies of a patient presentation is often available. This is *not* representative of the limited patient history to guide acute image ordering in the emergency room. To best simulate decision-making contexts with limited patient information available, we first needed to curate a dataset of patient scenario descriptions—or "one-liners"—and corresponding ground-truth labels. We call this resource the **RadCases Dataset** and detail its construction below.

To build the RadCases dataset, we leveraged five publicly available, retrospective sources of textual data. Firstly, we prompted the GPT-3.5 (gpt-3.5-turbo-0125) LLM from OpenAI to generate 16 **Synthetic** patient-cases with a chief complaint related to each of the 11 particular ACR AC Panels related to diagnostic radiology.

To include more challenging patient cases, we also introduced the Medbullets patient cases consisting of challenging United States Medical Licensing Examination (**USMLE**) Step 2- and 3-style cases introduced by Chen et al.[23] The original Medbullets dataset consisted of paragraph-form patient cases accompanied by a multiple-choice question; to convert each question to a patient one-liner, we used the first sentence of each patient case.

Similarly, we leveraged the **JAMA** Clinical Challenge and **NEJM** Case Record datasets that include challenging, real-world cases published in the Journal of the American Medical Association (JAMA) and the New England Journal of Medicine (NEJM), respectively. These patient cases are often described as atypical presentations of complex diseases that are noteworthy enough to be published as resources for the broader medical community. The JAMA Clinical Challenge (resp., NEJM Case Record) dataset was initially introduced by Chen et al.[23] (resp., Savage et al.[11]); we follow the same protocol as for the Medbullets dataset described above to programmatically convert these document-form patient cases into patient one-liners.

Finally, we sought to evaluate LLMs on patient summaries written by clinicians in a real-world emergency department. We constructed the **BIDMC** dataset from anonymized, de-identified patient discharge summaries introduced by Johnson et al.[24] in the MIMIC-IV dataset by taking the first sentence of each discharge summary as the patient one-liner. Briefly, the original MIMIC-IV dataset includes electronic health record data of patients from the Beth Israel Deaconess Medical Center (BIDMC) admitted to either the emergency department or an intensive care unit (ICU) between 2008 and 2019.[24] We restrict our constructed one-liner dataset to those from the discharge summaries of a subset of 100 representative patients.

A patient one-liner was excluded from evaluation if any of the following exclusionary criteria applied: (1) the ACR AC did not provide any guidance for the chief complaint (e.g., a primary dermatologic condition); (2) an appropriate imaging study was performed and/or a diagnosis was already made; (3) the one-liner did not include sufficient information about the patient; or (4) the one-liner did not refer to a specific patient presentation (e.g., one-liners extracted from epidemiology-related USMLE practice questions).

**Formulating a strategy for LLM evaluation using the RadCases dataset**
Our curated RadCases dataset consists of 1,599 patient one-liner scenarios constructed from five different sources representing a diverse panel of patient presentations and clinical scenarios: (1) RadCases-Synthetic; (2) RadCases-USMLE; (3) RadCases-JAMA; (4) RadCases-NEJM; and (5) RadCases-BIDMC.

Our next task was to annotate ground-truth labels to each of the patient scenarios in the RadCases dataset. An intuitive ground-truth label might be to assign a single "best" imaging study (or lack thereof) to order for each patient scenario. However, such a singular ground-truth label is often non-existent: imaging studies can vary largely by clinician preference—even amongst expert physicians[25–27]—and available hospital resources. Furthermore, the ultimate goal of the RadCases dataset is to align LLMs with evidence-based guidelines for image ordering; datums consisting solely of patient scenario-imaging study pairs arguably contain weak, implicit signals on the underlying guidelines that dictate the "best" imaging study.

In light of these challenges, we instead hypothesized that LLMs could yield better imaging study predictions if evidence-based guidelines were included as an explicit module in the patient scenario-imaging study inference pipeline. If a language model classified patient scenarios to a specific guideline (i.e., a Topic of the ACR AC), then the best imaging study would be deterministically identified by the content of the guideline itself. More concretely, we looked to query LLMs to map input one-liners to output ACR Appropriateness Criteria Topics, and then programmatically map these Topics to their corresponding evidence-based imaging recommendations (**Fig. 1a**). This strategy helped inform our efforts in assigning ground-truth labels to the RadCases dataset discussed below.

**Labelling one-liners by ACR Appropriateness Criteria topics**
In alignment with this plan, two fourth-year U.S. medical students—supervised by an attending radiologist—manually annotated all RadCases scenarios according to the ACR AC Topic that best describes the patient case. As an illustrative example, the input one-liner "49M with HTN, IDDM, HLD, and 20 pack-year smoking hx p/w 4 mo hx SOB and non-productive cough" is mapped to the ACR AC Topic "Lung Cancer Screening."

In scenarios where multiple ACR AC Topics might be applicable to a single patient case, the more acute, life-threatening scenario was used as the ground-truth label. Patient cases that were not well-described by any of the available ACR AC Topics were excluded from the dataset.

Neurologic topics were the most common label in all 5 RadCases subsets, followed by cardiac and gastrointestinal conditions (**Supplementary Fig. 1, Supplementary Table 1**). Of note, while there are 224 unique diagnostic imaging topics in the ACR AC, only 161 (71.9%) of all topics had nonzero support in the dataset. Furthermore, 73 (32.6%) unique topics are represented in the Synthetic dataset; 61 (27.2%) in the USMLE dataset; 119 (53.1%) in the JAMA dataset; 70 (31.3%) in the NEJM dataset; and 47 (21.0%) in the BIDMC dataset.

**Evaluating large language models on the RadCases dataset**
Using the annotated RadCases dataset, we evaluated 6 state-of-the-art, publicly available LLM models. (1) **DBRX** Instruct (databricks/dbrx-instruct) from Databricks is an open-source mixture-of-experts (MoE) model with 132B total parameters.[28] (2) **Llama 3** 70B Instruct (meta-llama/Meta-Llama-3-70B-Instruct) from Meta AI is an open-source LLM with 70B total parameters.[29] (3) **Mistral** 8x7B Instruct (mistralai/Mixtral-8x7B-Instruct-v0.1) from Mistral AI is an open-source sparse MoE model with 47B total parameters.[30] (4)

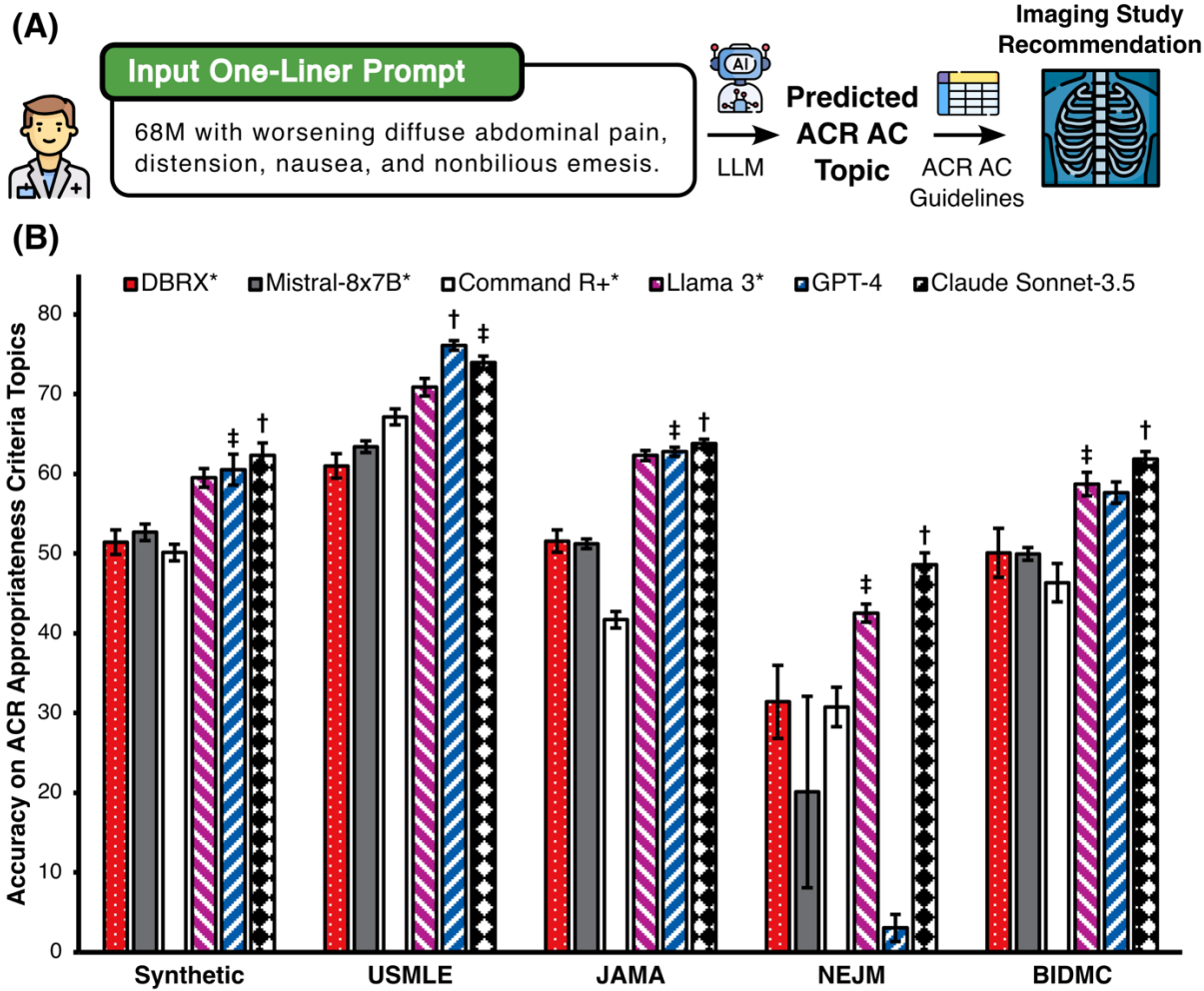

**Figure 1: Baseline LLM Performance on the RadCases Dataset.** **(A)** To align LLMs with the evidence-based ACR Appropriateness Criteria (AC), we query a language model to return the most relevant ACR AC Topic (224-way classification task) given an input patient one-liner description. We then programmatically query the ACR AC to deterministically return the most appropriate diagnostic imaging study (or lack thereof) given the predicted topic. **(B)** We evaluate six language models on their ability to correctly identify the ACR AC Topic most relevant to a patient one-liner. Open-source models are identified by an asterisk, and the best (second best) performing model for a RadCases dataset partition is identified by a dagger (double dagger). Error bars represent ± 95% CI over $n = 5$ independent experimental runs.

Command R+ (CohereForAI/c4ai-command-r-plus) from Cohere for AI is an open-source retrieval-optimized model with 104B total parameters.[31] (5) **GPT-4** Turbo (gpt-4-turbo-2024-04-09) from OpenAI and (6) **Claude Sonnet-3.5** (anthropic.claude-3-5-sonnet-20240620-v1:0) from Anthropic AI are proprietary LLMs with confidential model sizes.[32,33] Each of these six models were assessed in their ability to identify the singular best ACR AC Topic to an input patient one-liner from the RadCases dataset. To date, there are 224 possible Topic labels to choose from in the ACR AC.

**Figure 1b** shows the performance of the LLMs evaluated on each of the RadCases dataset subsets. Notably, Claude Sonnet-3.5 performs the best on 4 out of the 5 subsets (i.e., Synthetic, JAMA, NEJM, and BIDMC) and the second best on the remaining subset (i.e., USMLE). Furthermore, Claude Sonnet-3.5 outperformed all open-source models with statistical significance (two-sample, two-tailed homoscedastic *t*-test; Synthetic $p = 0.0037$; USMLE $p = 0.0003$; JAMA $p = 0.0016$; NEJM $p < 0.0001$; BIDMC $p = 0.0010$). Separately, Llama 3 outperformed all other evaluated open-

Yao MS et al. Aligning Language Models with the ACR Appropriateness Criteria. 4

| Model Error | Example Patient One-Liner | Model Predicted Label | Ground-Truth Label |
|---|---|---|---|
| **Hallucination of Patient Presentation**: The predicted topic is incorrect because the model assumes the presence of a sign or symptom that is not present in the input patient one-liner. | An otherwise healthy man in his 60s presented to a tertiary otolaryngology clinic reporting a 1-year medical history of progressively worsening hoarseness and dyspnea. | Hearing Loss and/or Vertigo | Chronic Dyspnea-Noncardiovascular Origin |
| | A teenage girl presented with new onset of bilateral upper extremity numbness and paresthesias. | Neck Mass/Adenopathy | Myelopathy |
| **Hallucination of ACR AC Topics**: The predicted topic is a relevant label to describe the patient case, but is not a documented ACR AC Topic. | A 77-year-old man presents to the emergency department with a complaint of sudden onset weakness in his right upper extremity. | Acute Stroke or TIA | Cerebrovascular Diseases-Stroke and Stroke-Related Conditions |
| | A woman in her 80s with a history of recurrent oral cancer presented with a new, biopsy-confirmed, squamous cell carcinoma (SCC). | Staging and Follow-up of Head and Neck Cancer | Staging and Post-Therapy Assessment of Head and Neck Cancer |
| | A 25-year-old woman presents to her primary care physician complaining of recent hair growth along her jawline, now requiring her to shave every 2 days. | Hirsutism | Clinically Suspected Adnexal Mass, No Acute Symptoms |
| **Patient Demographic**: The model incorrectly categorizes a pediatric topic label for an adult patient, or vice versa. | A 4-year-old boy comes to the pediatrician with his mother with a 3-day history of cough and runny nose with decreased oral intake over the past 24 hours. | Acute Respiratory Illness in Immunocompetent Patients | Pneumonia in the Immunocompetent Child |
| | A teenage girl newly diagnosed with idiopathic intracranial hypertension by her neurologist presented with 2 weeks of headaches, dizziness, and blurred vision. | Headache | Headache-Child |
| **Lack of Clinical Knowledge**: The model fails to recognize a constellation of clinical findings that is suggestive of a topic that is either more life-threatening or a better explanation of all presenting symptoms. | A 69-year-old man presents to clinic due to shortness of breath, worsening pain in his right shoulder, and episodes of hemoptysis. | Hemoptysis | Noninvasive Clinical Staging of Primary Lung Cancer |
| | A 62-year-old woman undergoing peritoneal dialysis (PD) for kidney failure due to IgA nephropathy presented to the PD clinic with a 1-day history of severe abdominal pain and cloudy PD fluid. | Acute Nonlocalized Abdominal Pain | Sepsis |
| | A woman in her late 50s presented to the clinic with severe right eye pain and a headache for the past 2 days. | Orbits, Vision, and Visual Loss | Headache |

**Table 1: Failure Modes of Large Language Models in ACR AC Topic Classification Prediction.** Qualitatively, we found that there were 4 common failure modes to explain the majority of incorrect predictions made by LLMs: (1) Hallucination of additional clinical signs or symptoms that were not described in the input patient scenario; (2) Hallucination of nonexistent (but often semantically related) ACR AC Topics; (3) assigning pediatric Topic categories to adult patients or vice versa; or (4) a lack of clinical knowledge to unify constellations of findings under a single likely diagnosis.

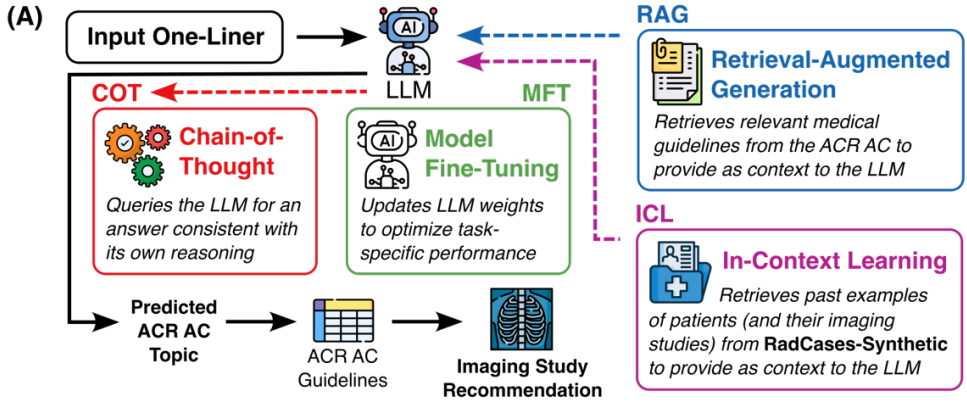

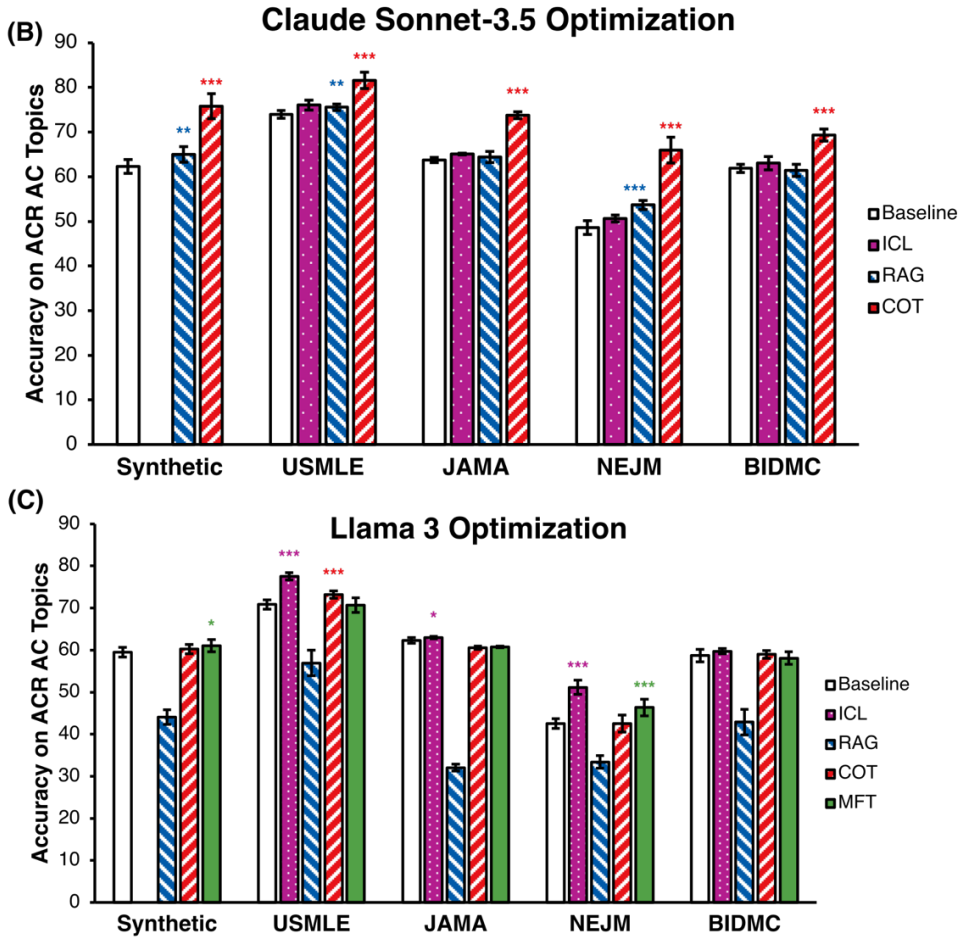

**Figure 2: Optimizing LLM Performance on the RadCases Dataset. (A)** We explore 4 strategies to further improve LLM alignment with the ACR AC: RAG and ICL provide additional context to an LLM as input, COT encourages deductive reasoning, and MFT optimizes the weights of the LLM itself. Each optimization strategy is independently implemented and compared against the baseline prompting results in **Figure 1** for **(B)** Claude Sonnet-3.5 and **(C)** Llama 3. Because ICL uses the RadCases-Synthetic dataset as its corpus of examples, we do not evaluate ICL on the RadCases-Synthetic subset to avoid data leakage. Error bars represent ± 95% CI over $n = 5$ independent experimental runs.

source models across all 5 RadCases subsets (two-sample, two-tailed homoscedastic *t*-test; *p* < 0.0002 for all 5 subsets). Based on these results, we chose to further optimize Claude Sonnet-3.5 and Llama 3 in subsequent experiments as the most promising overall and open-source large language models, respectively. Common failure modes to explain incorrect model predictions by all 6 LLMs are detailed in **Table 1**, and additional classification metrics are described in **Supp. Fig. 2**.

**Optimizing large language models for imaging ordering in acute clinical workflows**
While Claude Sonnet-3.5 and Llama 3 demonstrated impressive baseline accuracy on the RadCases dataset, recent work have introduced techniques to improve the performance of generative language models. For example, retrieval-augmented generation (RAG) provides relevant context to language models retrieved from an information corpus (i.e., the ACR AC narrative medical guidelines written by expert radiologists) to help improve the generative process. In-context learning (ICL) provides relevant examples of patient one-liners and their corresponding topic labels (i.e., examples from the RadCases-Synthetic dataset) as relevant context to improve the zero-shot performance of language models. Chain-of-thought (COT) prompting is a strategy to improve the complex reasoning abilities of language models by encouraging sequential, logical steps to arrive at a final answer. Finally, model fine-tuning (MFT) directly updates the parameters of a language model to improve its performance on a specific task. We assess the all four strategies using Llama 3, and the zero-shot strategies RAG, ICL, and COT using Claude Sonnet-3.5 as there is no publicly available application programming interface (API) to fine-tune the proprietary model to date (**Fig. 2a**).

**Figure 2b** shows that COT (chain-of-thought prompting) is the most effective strategy for Claude Sonnet-3.5, resulting in improvements of up to 17% in ACR AC Topic classification accuracy and consistent improvements across all five RadCases dataset subsets (two-sample, one-tailed, homoscedastic *t*-test; *p* < 0.0001 for all subsets). Interestingly, this same strategy does not translate well to Llama 3 (**Fig. 2c**); COT marginally improves upon baseline prompting for Llama 3 only on the USMLE RadCases dataset. Instead, ICL (in-context learning) was the most effective prompt engineering strategy for Llama 3, resulting in improvements of up to 9% on ACR AC Topic classification accuracy compared with naïve prompting (two-sample, one-tailed homoscedastic *t*-test; *p* < 0.0001 for USMLE and NEJM datasets). Additional fine-grained optimization results are included in **Supplementary Figs. 3-7**.

Our results show that while prompt engineering and other optimization techniques can indeed be effective in improving the performance of different language models on this task, the trends in improvements can be LLM-specific and fail to generalize across different language models. This finding highlights the inherit challenge in optimizing such models for challenging tasks such as diagnostic image ordering via alignment with the ACR AC.

**Validating the LLM prediction pipeline**
So far, our core hypothesis in engineering LLMs for diagnostic image ordering is that using language models to map patient scenarios to ACR AC Topics instead of directly to imaging studies will improve their alignment with the ACR AC. In **Figure 1b**, we demonstrated that LLMs could achieve reasonable accuracy on this Topic classification task; in **Figure 2b-c**, we further optimized two state-of-the-art language models using prompt engineering techniques and model fine-tuning. Based on these results, we now sought to validate our original hypothesis and evaluate whether assigning ACR AC Topic predictions to patient one-liners can meaningfully improve LLM performance in diagnostic image study ordering.

We first mapped the ground-truth ACR AC Topic labels in the RadCases dataset to the ground-truth imaging study recommended by the relevant Topic guidelines. We analyzed 3 different LLM inference pipelines using Claude Sonnet-3.5: (1) **Baseline**, which queries an LLM to directly recommend a diagnostic imaging study; (2) **Evidence-Based Baseline**, which queries an LLM to recommend a ACR AC Topic that is then mapped to the imaging study; and (3) **Evidence-Based Optimized**, which is the same as (2) but uses the optimized COT prompting strategy from **Figure 2b** for Claude Sonnet-3.5 (**Fig. 3a**).

Our results demonstrate that our evidence-based algorithm of leveraging LLMs to map to ACR AC Topics provides significant improvements in the overall imaging accuracy achieved by the model. Across all 5 RadCases dataset subsets, our Evidence-Based Baseline (resp., Evidence-Based Optimized) pipeline outperforms the Baseline pipeline by at least 40% (resp., 42%) on imaging accuracy (two-sample, one-tailed homoscedastic $t$-test; $p < 0.0001$) (**Fig. 3b**).

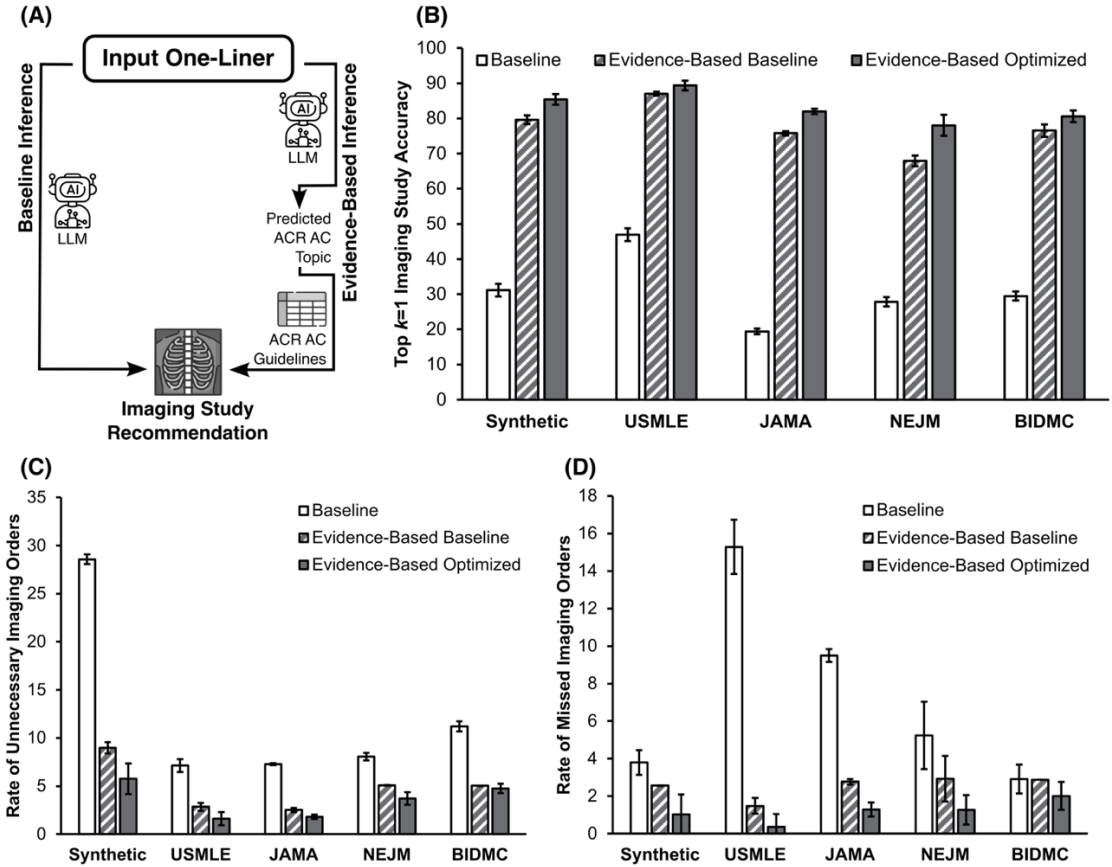

**Figure 3: Comparison of Baseline and Evidence-Based Inference Pipelines with Claude Sonnet-3.5.** **(A)** Using our evidence-based inference pipeline, we query the LLM to predict the single ($k = 1$) ACR AC Topic most relevant to an input patient one-liner, and programmatically refer to the evidence-based ACR AC guidelines to make the final recommendation for diagnostic imaging. An alternative approach is the baseline inference pipeline where we query the LLM to recommend a diagnostic imaging study directly without the use of the ACR AC. **(B)** Our evidence-based pipelines (both using baseline prompting and optimized using chain-of-thought (COT) prompting) significantly outperform the baseline pipeline by up to 62.6% (two-sample, one-tailed, homoscedastic $t$-test; $p < 0.0001$ for all RadCases datasets). At the same time, the also reduce the rates of both **(C)** unnecessary imaging orders and **(D)** missed imaging orders (two-sample, one-tailed, homoscedastic $t$-test; $p < 0.05$ for all RadCases datasets). Error bars represent ± 95% CI over $n = 5$ independent experimental runs.

Interestingly, while the Evidence-Based Optimized pipeline significantly outperformed the Baseline pipeline on ACR AC Topic classification accuracy (**Fig. 2b**), we did not observe a statistically significant improvement in the optimized versus baseline Evidence-Based pipelines on the imaging classification accuracy (two-sample, one-tailed homoscedastic $t$-test; $p \geq 0.346$ for each of the 4 RadCases dataset subsets).

Qualitatively, we found that although the Evidence-Based Baseline pipeline achieved a lower ACR AC Topic classification accuracy compared to the Evidence-Based Optimized inference strategy, its incorrect predictions were still closely related to the correct answer and underlying patient pathology. For example, a ground truth ACR AC Topic might be "Major Blunt Trauma" and the LLM prediction "Penetrating Torso Trauma;" although the LLM identified the incorrect ACR AC Topic label, both Topics warrant a "Radiography trauma series." As a result, both the optimized and baseline Evidence-Based pipelines achieve comparable imaging accuracy, and significantly improve upon the Baseline pipeline.

We are also interested in the false positive and false negative rates in image ordering. Formally, false positives are cases where an imaging study is unnecessarily ordered, and false negatives are cases where a diagnostic imaging study was warranted but not ordered. Both Evidence-Based pipelines again outperform the Baseline pipeline according to both metrics, significantly reducing the rates of false positives and false negatives (two-sample, one-tailed homoscedastic *t*-test; $p < 0.0001$ for the Synthetic, USMLE, JAMA, and NEJM subsets).

**Investigating autonomous image ordering using LLMs versus standard of care**

Based on the initial results in **Figure 3** and **Supplementary Figure 8**, we next looked to assess if state-of-the-art, optimized language models could be used to accurately order imaging studies for acutely presenting patients without clinician intervention. In a retrospective study, we extracted a diverse sample of 242 anonymized, de-identified discharge summaries derived from the MIMIC-IV dataset.[24] These discharge summaries were taken from the medical records of 100 real patient admissions between 2008-2019 from the Beth Israel Deaconess Medical Center (Boston, MA).[24] To account for the limited patient information available in acute presentations, we manually truncated the discharge summaries to only include relevant patient history and vitals (**Supplementary Figure 9**). Discharge summaries were excluded from our analysis if either (1) the ACR Appropriateness Criteria contained no evidence-based guidance relevant to the patient scenario; or (2) the scenario described a patient admission that was not made in the emergency department (e.g., ICU downgrade to hospital floor). A total of 141 final patient scenarios were included in our analysis.

Using these patient scenarios, we prompted language models to predict up to three imaging studies that would be appropriate to order for a given patient. We evaluated two LLMs from our original RadCases evaluation suite: Claude Sonnet-3.5 from Anthropic AI using chain-of-thought (COT) prompting, and Llama-3 70B Instruct from Meta AI using no special prompt engineering. We chose to evaluate these two models because they were the best performing proprietary and open-source models on the RadCases benchmarks, respectively (**Fig. 1b**). Separately, we manually analyzed each of the full, original discharge summaries to determine what imaging study(s) where ordered by the patient's physician. The imaging studies ordered by both clinicians and language models were compared against the ground-truth best imaging study(s) as determined by consensus between an expert abdominal radiologist and two fourth-year U.S. medical students at the University of Pennsylvania.

Our results suggest that autonomous LLMs achieve comparable accuracy to that of clinicians in ordering diagnostic imaging (**Fig. 4a**): there was no statistically significant difference between the performance of Claude Sonnet-3.5 (accuracy 40.8%) and clinicians (accuracy 39.0%) (McNemar test; $\chi^2 = 2.63$; df = 1; $p = 0.105$). Similarly, there was no statistically significant difference between Llama 3 (accuracy 33.3%) and clinicians (McNemar test; $\chi^2 = 0.327$; df = 1; $p = 0.568$). Across the patient cases assessed, clinicians ordered an average of 1.40 (95% CI: [1.28 – 1.52]) imaging studies per case; in contrast, Claude Sonnet-3.5 ordered an average of 1.74 (95% CI: [1.59 – 1.88]) and Llama 3 and average of 2.00 (95% CI: [1.84 – 2.16]) studies per case (**Fig. 4d**). Both language models assessed ordered more imaging studies than their clinician counterparts with statistical significance (two-sample paired *t*-test; Claude Sonnet-3.5: $p = 0.0006$; Llama 3: $p < 0.0001$).

We also evaluated the rates of both unnecessary and missed imaging studies (**Fig. 4b-c**). Claude Sonnet-3.5 was non-inferior to clinicians according to both metrics, achieving a false positive rate (FPR) of 6.90% (clinician FPR = 5.49%) (McNemar test; $\chi^2 = 0.000$; df = 1; $p = 1.00$) and false negative rate (FNR) of 13.7% (clinician FNR = 5.98%) (McNemar test; $\chi^2 = 3.37$; df = 1; $p = 0.066$). Llama 3 demonstrated an FPR of 8.12% with no statistically significant difference compared to that of clinicians (McNemar test; $\chi^2 = 0.500$; df = 1; $p = 0.480$); however, Llama 3 suffered from a significantly higher FNR of 22.2% (McNemar test; $\chi^2 = 10.45$; df = 1; $p = 0.0012$). Altogether, these results suggest that LLMs are promising tools for image ordering in clinical workflows, with proprietary models like Claude Sonnet-3.5 outperforming their open-source counterparts.

Finally, to gauge the similarity between recommendations made by different language models and clinicians, we computed the pairwise Dice-Sørensen coefficient (DSC) between imaging recommendations made by different decision makers (**Fig. 4e**). According to this metric, we found that recommendations made by different language models aligned significantly more closely than those from language models and clinicians: the DSC between (1) Claude Sonnet-3.5 and Llama 3 was 0.517 (95% CI: [0.449 – 0.584]); (2) Claude Sonnet-3.5 and clinicians was 0.126 (95% CI: [0.073 – 0.178]); and (3) Llama 3 and clinicians was 0.091 (95% CI: [0.049 – 0.133]).

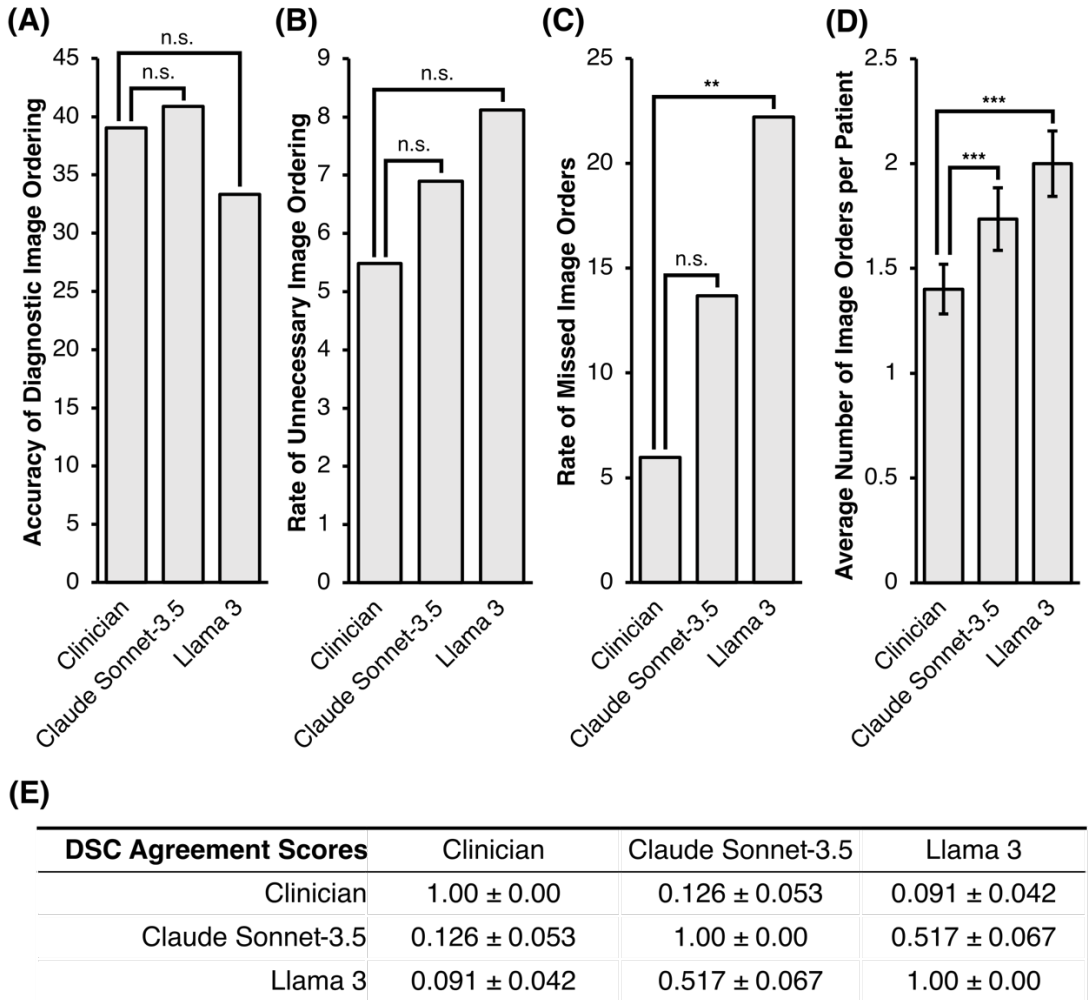

**Figure 4: Retrospective Study of Clinician-Ordered versus LLM-Ordered Imaging Studies. (A)** We compare the diagnostic imaging studies ordered by the prompt-optimized LLMs Claude Sonnet-3.5 and Llama 3 against those ordered by clinicians in a retrospective study. Claude Sonnet-3.5 and Llama 3 achieve **(A)** accuracy scores and **(B)** false positive rates that are not significantly different from those achieved by clinicians, and Claude Sonnet-3.5 achieves a **(C)** false negative rate that is not significantly different from clinicians. **(D)** However, both language models assessed order significantly more imaging studies per patient than clinicians. **(E)** According to the Dice-Sørensen Coefficient (DSC) metric, Claude Sonnet-3.5 and Llama 3 order imaging studies that are more similar to one another than to clinicians (two sample, two-tailed homoscedastic *t*-test; $p < 0.0001$).

**Evaluating language models as support tools for clinician diagnostic image ordering**
In our experiments above, we assessed LLMs as autonomous agents for clinical decision making. Such retrospective studies help clarify the technical capabilities and limitations of these models compared with standard of care. However, LLMs could also act as *assistants* for clinicians in diagnostic image ordering.

To evaluate the utility of our evidence-based LLMs as clinical assistants, we conducted a prospective randomized control trial[i] asking volunteered clinician participants to order diagnostic imaging studies for simulated patient scenarios in an online testing environment. Participants were U.S. medical students and emergency medicine resident physicians recruited from the Perelman School of Medicine and the Hospital of the University of Pennsylvania. This study was exempted by the University of Pennsylvania Institutional Review Board (Protocol #856530).

Each study participant was asked to order a single imaging study (or forego imaging if not indicated) for 50 simulated patient cases. For each participant, a random 50% of the patient cases included recommendations generated by Claude Sonnet-3.5 using the evidence-based optimized inference strategy in **Fig. 3**. To simulate the acuity and high-pressure of many emergency room environments, participants were required to complete the study at an average rate of 1 case per minute in a single setting. We then fitted a regression model according to

$$y_{s,q} = \beta_0 + (\beta_1 * \text{WithLLMGuidance}_{s,q}) + \theta_q + \chi_s + \varepsilon_{s,q} \quad (1)$$

where $s$ indexes study participants and $q$ study questions, and $y_{s,q}$ is a binary variable indicating whether participant $s$ answered study question $q$ correctly. Here, $\theta$ is a $q$-vector of study question fixed effects, $\chi_s$ are control variables specific to the study participant (i.e., whether the study participant is a physician or medical student, the participant's personal experience with AI, and the participants sentiment regarding AI), and $\varepsilon_{s,q}$ is the error term. We estimate **Equation 1** using standard errors clustered at the study participant level and question level. Furthermore, $\text{WithLLMGuidance}_{s,q}$ is a binary indicator that indicates whether LLM-generated guidance was provided for question $q$ for participant $s$, respectively.

Study participants generally found the study task challenging, with an average accuracy of 15.8% (95% CI: [12.2% – 19.3%]) without LLM guidance and 25.0% (95% CI: [20.7% – 29.3%]) with guidance. Offering LLM-based recommendations using our evidence-based optimized pipeline improved image ordering accuracy with statistical significance ($\beta_1$ = 0.081; 95% CI: [0.022 – 0.140]; $p$ = 0.011) for both medical students and resident physicians.

To verify that participants were indeed taking advantage of LLM-generated recommendations when made available, we fitted a separate regression model analogous to that in **Equation 1** that instead measures that binary agreement between LLM recommendations and participant answers as the dependent variable. As expected, the agreement between answers and assistant recommendations increases when the recommendations are made available to the clinician ($\beta_1$ = 0.141; 95% CI: [0.050 – 0.233]; $p$ = 0.005). These results suggest that language models can act as clinical assistants to help clinicians order imaging studies more aligned with evidence-based guidelines.

Similarly, we did not observe statistically significant differences in either the false positive rate ($\beta_1$ = 0.008; 95% CI: [-0.012 – 0.027]; $p$ = 0.418) or false negative rate ($\beta_1$ = -0.019; 95% CI: [-0.068 – 0.030]; $p$ = 0.431). This ensures that the improvements in accuracy scores with LLM guidance without significantly increasing the number of unnecessary or missed imaging studies ordered by clinicians. Additional analysis is included in **Supplementary Tables 5-8**, and discussion of experimental results in **Appendix A**.

## DISCUSSION

Our study investigates the potential of LLMs in the domain of diagnostic image ordering—a task critical to the timely and appropriate management of acute patient presentations. In our work, we introduce the first publicly available dataset of patient one-liners and ACR AC Topics, and both evaluate and optimize state-of-the-art LLMs to map patient scenarios to appropriate imaging studies. To interrogate the future clinical applicability of these AI tools, we conduct both retrospective and prospective studies that demonstrate that LLMs such as Claude Sonnet-3.5 can be valuable tools for clinical decision support in acute care settings.

Importantly, we demonstrate how integrating evidence-based guidelines (i.e., the ACR AC) directly into the LLM-based inference pipeline can significantly improve the accuracy of clinical recommendations. This approach not only aligns model predictions with established guidelines, but also provides a robust framework for reducing the rates of both unnecessary and missed imaging orders. In theory, such a framework could be readily translatable for other clinical use cases that make use of available guidelines, such as the American College of Gastroenterology guidelines to determine clinical indications for endoscopy,[34,35] or the American Society of Addiction Medicine guidelines for the management of alcohol withdrawal syndrome.[36] We leave these potential future applications of LLM-based CDS tools for future work.

---

[i] Pre-registration on AsPredicted, #185312. Link:
https://aspredicted.org/x6b9-rcgh.pdf

We also highlight the challenges associated with integrating novel LLM toolkits into existing clinical workflows. In our prospective study, we found that the utility of LLM clinical assistants can be largely dependent on factors such as user expertise, acuity of care, and existing user attitudes on AI. Careful consideration of these factors and others are crucial to ensure that LLMs are used responsibly and can improve patient care.

Of note, our results consistently demonstrate that proprietary language models, such as Claude Sonnet-3.5, consistently and significantly outperform open-source models. While the performance of Claude Sonnet-3.5 is indeed impressive, it is unlikely that current publicly available inference APIs for the model are sufficient for widespread clinical deployment, as many hospitals understandably express concerns over patient privacy and unknown data handling practices from third-party vendors. A viable strategy that leverages future models with comparable performance to existing proprietary models and is simultaneously compliant with healthcare regulations and best practices remains an important bottleneck for the use of LLMs as clinical decision support tools.

This study also has its limitations. Firstly, while constructing the RadCases dataset used to evaluate LLMs, we excluded cases where there exist no ACR AC guidelines relevant to the patient scenario. Such cases might include primary dermatologic conditions and cases where insufficient ACR evidence exist. While this strategy allowed us to primarily focus on how existing LLMs perform according to the ACR AC, future work is warranted to investigate how these models handle scenarios where there is a lack of relevant guidelines. It is also important to ensure that LLM-based tools operate equitably across diverse patient populations and clinical scenarios before they are deployed across hospitals systems. Future studies are warranted to investigate potential biases in model predictions—and develop strategies to mitigate them.

In conclusion, our study highlights the potential of LLMs to enhance the process of diagnostic image ordering by leveraging evidence-based guidelines. By mapping patient scenarios to ACR AC Topics and using optimized model strategies, LLMs can improve accuracy and efficiency in imaging decisions. Our findings suggest that LLMs could play a transformative role in supporting clinicians and improving patient care in acute diagnostic workflows.

## MATERIALS AND METHODS
### Optimization of zero-shot prompt engineering and fine-tuning methods

In **Figure 2**, we explore 4 distinct LLM optimization strategies—retrieval-augmented generation (RAG), in-context learning (ICL), chain-of-thought (COT) prompting, and model fine-tuning (MFT)—to improve the ability of LLMs like Claude Sonnet-3.5 and Llama 3 to accurately predict relevant ACR AC Topics from input patient one-liner scenarios. All LLM query prompts are included in **Supplementary Table 5**. For all experiments described herein, LLM prompts were first optimized on a small, holdout set of 10 synthetically generated one-liners that were not a part of the RadCases dataset before being used for all experiments reported herein.

In our RAG approach, we first constructed the relevant reference corpus of guidelines made publicly available by the American College of Radiology (ACR). A link to our custom script implementation is made publicly available at [this URL](). Using a custom Python script included in our publicly available code, we first used a web scraper, in compliance with the ACR Terms and Conditions, to download relevant Portable Document Format (PDF) narrative files from [acsearch.acr.org/list]() on July 17, 2024. Each ACR AC Topic features one accompanying narrative document, resulting in a total of 224 narrative files extracted. We then used the [Unstructured IO open-source library]() to extract the PDF content into raw text, and chunked the text into 3,380 disjoint corpus documents with the number of characters ranging between 1,119 and 2,048 characters per document. Our strategy for constructing the retrieval corpus is identical to that used by Xiong et al.[37]

Using this corpus of relevant guidelines written by the ACR, we explored 8 different retriever algorithms to use for RAG: (1) **Random**, which randomly retrieves *k* documents from the corpus over a uniform probability distribution; (2) Okapi **BM25** bag-of-words retriever[38]; (3) **BERT**[39] and (4)

**MPNet**[40] trained on unlabeled, natural language text; (5) **RadBERT**[13] from fine-tuning BERT on radiology text reports; (6) MedCPT[41] leveraging a transformer trained on PubMed search logs; and (7) **OpenAI** (text-embedding-3-large) and (8) **Cohere** (cohere.embed-english-v3) embedding models from OpenAI and Cohere for AI, respectively. Retrievers (3) – (8) are embeddings-based retrievers that leverage cosine similarity as the ranking function. These 8 retrievers represent a diverse array of novel, well-studied, domain-agnostic, and domain-specific retrievers for RAG applications. In **Figure 2b-c**, we report the results using the best retriever specific to each language model and RadCases dataset subset, fixing the number of retrieved documents to $k = 8$ for each retriever. We include the experimental results for each individual retriever in **Supplementary Fig. 3**.

Separately in our ICL approach, we use the RadCases-Synthetic dataset partition as the corpus of examples to retrieve from, and experimentally validate the same 8 retrievers used in RAG for retrieving relevant one-liner/ACR AC Topic pairs to provide as context to the language model. In Figure 2b-c, we report the results using the best retriever specific to each language model and RadCases dataset subset, fixing the number of retrieved examples to $k = 4$ for each retriever. Because the RadCases-Synthetic dataset is used as the corpus, we do not evaluate ICL on the RadCases-Synthetic dataset as well to avoid data leakage. We include the experimental results for each individual retriever in **Supplementary Fig. 4**, and explore the effect of different values of $k$ (i.e., the number of retrieved examples) in **Supplementary Fig. 5**.

In COT prompting, we explore four different reasoning strategies identical to those employed by Savage et al.[11]: (1) **Default** reasoning, which does not specify any particular reasoning strategy for the LLM to use; (2) **Differential** diagnosis reasoning, which encourages the model to reason through a differential diagnosis to arrive at a final prediction; (3) **Bayesian** reasoning, which encourages the model to approximate Bayesian posterior updates over the space of ACR AC Topics based on the clinical patient presentation; and (4) **Analytic** reasoning, which encourages the model to reason through the pathophysiology of the underlying disease process. We include the experimental results for each individual reasoning strategy in **Supplementary Fig. 6**. In **Figure 2b-c**, we report the results using the best COT reasoning strategy specific to each language model and RadCases dataset subset. In **Figures 3-5**, we report results using the **Default** reasoning strategy when COT is leveraged together with Claude Sonnet-3.5.

For MFT, we explore three different fine-tuning strategies: (1) **Full** fine-tuning where all the parameters of the LLM are updated; and (2) Low-Rank Adaptation (**LoRA**)[42] and (3) Quantized Low-Rank Adaptation (**QLoRA**)[43] fine-tuning where only a subset of the LLM parameters are updated. We fix the number of training epochs to 3 and the learning rate to 0.0001. For LoRA (resp., QLoRA), we use a rank of 64 (resp., 512) and an $\alpha$ scaling value of 8 (resp., 8). Due to limitations on local compute availability, we were only able to run the QLoRA fine-tuning experiments on the internal experimental cluster; LoRA and Full fine-tuning experiments were performed using a third-party platform ([Together AI](Together AI)). Finally, we also investigate two different fine-tuning datasets for each of the three strategies: (1) fine-tuning on the RadCases-Synthetic dataset; and (2) fine-tuning on 250 cases where 50 random cases come from each of the five RadCases dataset subsets. To prevent data leakage, we do not evaluate fine-tuned models on the RadCases-Synthetic dataset in strategy (1), and avoid evaluation on any cases from the individual patients represented in the fine-tuning dataset in strategy (2). In **Figure 2c**, we report the results using the LoRA fine-tuning strategy and "mixed" fine-tuning dataset of 250 cases described above, as this led to consistently superior fine-tuning results across all datasets and language models that were evaluated. We report additional experimental results in **Supplementary Fig. 7**.

**Translating ACR AC Topics into imaging study recommendations**

In **Figure 3a**, we overview our Evidence-Based inference pipeline where we leverage LLMs to assign ACR AC Topics to input patient one-liner scenarios, and then deterministically map Topics to appropriate imaging studies based on the Appropriateness Criteria guidelines. These LLM-generated recommendations were used as the basis of our retrospective and prospective studies described in our work. Determining this mapping of Topics to imaging studies is a non-trivial task: for any particular Topic, there are often multiple, nuanced clinical variants that are described the ACR AC. For example, for the "Suspected Pulmonary Embolism" Topic, there are 4 variants in the guidelines as of August 2024:

1. Suspected pulmonary embolism. Low or intermediate pretest probability with a negative D-dimer. Initial imaging.
2. Suspected pulmonary embolism. Low or intermediate pretest probability with a positive D-dimer. Initial imaging.
3. Suspected pulmonary embolism. High pretest probability. Initial imaging.
4. Suspected pulmonary embolism. Pregnant patient. Initial imaging.

Each of these variants have different imaging recommendations: for example, variant (1) does not warrant any imaging study according to the ACR AC, whereas both computed tomography angiography (CTA) pulmonary arteries with intravenous (IV) contrast and a ventilation-perfusion (V/Q) scan lung are appropriate studies for variant (3). To define a deterministic mapping of topics to imaging studies, we therefore needed to isolate a single variant for each topic.

Our research team manually parsed through each of the 224 Topics to determine this single variant. In general, the process involved reverse engineering a "typical" patient presentation that would be described by a given Topic. In the above example, we reasoned that an acutely presenting patient where the most relevant Topic is "Suspected Pulmonary Embolism" would likely have a high pretest probability for a pulmonary embolism. Furthermore, pregnant patients are less common than non-pregnant patients in the emergency room, and the appropriate imaging studies for variant (3) are also appropriate for variant (4). For this reason, variant (3) was kept and the rest were discarded. As a result, a predicted imaging study of either CTA pulmonary arteries with IV contrast or V/Q scan lung were both considered correct answers in this example. If no imaging study was considered appropriate according to a guideline, then the ground-truth label was defined as "None."

**Participant recruitment and compensation**
In our work, we conducted a randomized clinical trial with senior U.S. medical students and U.S. emergency medicine physicians to evaluate if LLMs can serve as helpful assistants in deciding what imaging studies to order. Participants for this prospective study were recruited from the Perelman School of Medicine and the Hospital of the University of Pennsylvania where this study was conducted. We provided a monetary incentive of $50 USD to each opt-in, volunteer study participant, and the top 50% most accurate medical students and physicians (scored separately) within each treatment arm were compensated with an additional $10 USD. A total of 23 medical students and 7 resident physicians participated in our experiment; all participating medical students were required to have passed and completed the emergency medicine clinical rotation at the University of Pennsylvania to participate in this study.

**Participant task in prospective study**
Study participants were tasked with each ordering up to 1 diagnostic imaging study for a standardized set of 50 simulated patient case descriptions derived from the MIMIC-IV dataset.[24] Each case was presented on a custom-built website interface to display each one patient case at a time; a visual of the custom interface is shown in **Supplementary Figure 9**. For each case, participants selected an imaging study from a dropdown menu containing an alphabetized list of all 1150 diagnostic imaging studies officially recognized in the ACR Appropriateness Criteria. Of the 50 simulated cases, a random subset of 25 cases chosen at the per-participant level also showed LLM-generated recommendations for the participant to consult. Study participants were allowed to consult any online resources that they would typically use in evaluating patients in the emergency department, but were not allowed to consult any other individuals for assistance. In some simulated patient cases, more than one correct answer may be possible – study participants were instructed to select just one of those possible answers in these cases.

Separately, study participants were also asked to complete a 5-question multiple-choice survey asking questions about their prior experience with AI tools, and overall sentiment about the use of AI in medicine (**Supplementary Table 9**). All study participant answers to this short survey and the overall prospective study were anonymized and aggregated before analysis; participants were informed of this anonymization strategy in the informed consent.

**Experimental evaluation and statistical analysis**
All models and prompting techniques were evaluated on a single internal cluster with 8 NVIDIA RTX A6000 GPUs. The temperature of all models was set to 0 to minimize variability in the model outputs. Each experiment was run using 5 random seeds, and we computed the mean accuracy of each method with 95% confidence intervals (CIs) against the human-annotated ground truth labels. A $p$-value of $p < 0.05$ was used as the threshold for statistical significance. In all figures, "n.s." represents not significant (i.e., $p \geq 0.05$); a single asterisk $p < 0.05$; double asterisks $p < 0.01$, and triple asterisks $p < 0.001$. All statistical analyses were performed using Python software, version 3.10.13 (Python Software Foundation), the SciPy package, version 1.14.0 (Enthought),[44] and the PyFixest package, version 0.24.2.[45]

## DATA AVAILABILITY
All data are available within the article, supplementary information or the source data file provided with this paper. Source data are provided with this paper.

## CODE AVAILABILITY
Custom code used for large language model evaluation and prospective user studies is made publicly available at https://github.com/michael-s-yao/radGPT.

## ACKNOWLEDGEMENTS
The authors thank Kevin Johnson, M. Dylan Tisdall, and Mark Yatskar (listed alphabetically) at the University of Pennsylvania for their helpful discussions. We also appreciate the anonymous study participants and Wilma Chan, Lauren Conlon, Michael Abboud, Matthew Magda, and Mira Mamtani in the Department of Emergency Medicine at the University of Pennsylvania for their help and support in conducting the prospective clinician-AI study. This research was funded by the National Science Foundation Division of Computing and Communication Foundations (NSF Award CCF-1917852). M.S.Y. was supported by the NIH (F30 MD020264). A.C. was supported by the 2023 Alpha Omega Alpha Carolyn L. Kuckein Student Research Fellowship, and a research grant from the Department of Radiology at the University of Pennsylvania. J.C.G. was supported by the NIH (R01 EB031722). H.S. was supported by the Institute for Translational Medicine and Therapeutics' (ITMAT) Transdisciplinary Program in Translational Medicine and Therapeutics, and by the National Center for Advancing Translational Sciences of the National Institutes of Health under Award Number UL1TR001878. O.B. was supported by NSF Award CCF-1917852. The content is solely the responsibility of the authors and does not necessarily represent the official views of the NIH or the NSF.

# REFERENCES

1. Kwee RM, Toxopeus R, Kwee TC. Imaging overuse in the emergency department: The view of radiologists and emergency physicians. *European Journal of Radiology*. 2024;176:111536. doi:10.1016/j.ejrad.2024.111536

2. Baloescu C. Diagnostic imaging in emergency medicine: How much is too much? *Annals of Emergency Medicine*. 2018;72(6):637-643. doi:101016/j.annemergmed.2018.06.034

3. Salerno S, Laghi A, Cantone MC, Sartori P, Pinto A, Frija G. Overdiagnosis and overimaging: An ethical issue for radiological protection. *Radiol Med*. 2019;124(8):714-720. doi:10.1007/s11547-019-01029-5

4. Litkowski PE, Smetana GW, Zeidel ML, Blanchard MS. Curbing the urge to image. *Am J Med*. 2016;129(10):1131-1135. doi:10.1016/j.amjmed.2016.06.020

5. Bresnahan BW. Economic evaluation in radiology: Reviewing the literature and examples in oncology. *Acad Radiol*. 2010;17(9):1090-1095. doi:10.1016/j.acra.2010.05.020

6. American College of Radiology. ACR Appropriateness Criteria. Accessed June 30, 2024. https://acsearch.acr.org/list

7. Bautista AB, Burgos A, Nickel BJ, Yoon JJ, Tilara AA, Amorosa JK. Do clinicians use the American College of Radiology Appropriateness Criteria in the management of their patients? *Am J Roentgenol*. 2009;192(6):1581-1585. doi:10.2214/AJR.08.1622

8. Taragin BH, Feng L, Ruzal-Shapiro C. Online radiology appropriateness survey: Results and conclusions from an academic internal medicine residency. *Acad Radiol*. 2003;10(7):781-785. doi:10.1016/s1076-6332(03)80123-x

9. Nazario-Johnson L, Zaki HA, Tung GA. Use of large language models to predict neuroimaging. *J Am Coll Radiol*. 2023;20(10):1004-1009. doi:10.1016/j.jacr.2023.06.008

10. Zaki HA, Aoun A, Munshi S, Abdel-Megid H, Nazario-Johnson L, Ahn SH. The application of large language models for radiologic decision making. *J Am Coll Radiol*. 2024;21(7):1072-1078. doi:10.1016/j.jacr.2024.01.007

11. Savage T, Nayak A, Gallo R, Rangan E, Chen JH. Diagnostic reasoning prompts reveal the potential for large language model interpretability in medicine. *NPJ Digit Med*. 2024;7(20). doi:10.1038/s41746-024-01010-1

12. Chambon P, Cook TS, Langlotz CP. Improved fine-tuning of in-domain transformer model for inferring COVID-19 presence in multi-institutional radiology reports. *J Digit Imaging*. 2023;36(1):164-177. doi:10.1007/s10278-022-00714-8

13. Yan A, McAuley J, Lu X, et al. RadBERT: Adapting transformer-based language models to radiology. *Radiol Artif Intell*. 2022;4(4):e210258. doi:10.1148/ryai.210258

14. Tay SB, Low GH, Wong GJE, et al. Use of natural language processing to infer sites of metastatic disease from radiology reports at scale. *JCO Clin Cancer Inform*. 2024;8:e2300122. doi:10.1200/CCI.23.00122

15. Clusmann J, Kolbinger FR, Muti HS, et al. The future landscape of large language models in medicine. *Commun Med (Lond)*. 2023;3(1):141. doi:10.1038/s43856-023-00370-1

16. Malaviya C, Agrawal P, Ganchev K, et al. DOLOMITES: Domain-specific long-form methodical tasks. Published online 2024. doi:10.48550/arXiv.2405.05938

17. Ong JCL, Chang SYH, William W, et al. Ethical and regulatory challenges of large language models in medicine. *Lancet Digit Health*. 2024;6(6):E428-32. doi:10.1016/S2589-7500(24)00061-X

18. Omiye JA, Lester JC, Spichak S, Rotemberg V, Daneshjou R. Large language models propagate race-based medicine. *NPJ Digit Med*. 2023;6(195). doi:10.1038/s41746-023-00939-z

19. Allen B, Agarwal S, Coombs L, Wald C, Dreyer K. 2020 ACR Data Science Institute Artificial Intelligence Survey. *J Am Coll Radiol*. 2021;18(8):1153-1159. doi:10.1016/j.jacr.2021.04.002

20. Spotnitz M, Idnay B, Gordon ER, et al. A survey of clinicians' views of the utility of large language models. *Appl Clin Inform*. 2024;15(2):306-312. doi:10.1055/a-2281-7092

21. Kresevic S, Giuffrè M, Ajcevic M, Accardo A, Crocè LS, Shung DL. Optimization of hepatological clinical guidelines interpretation by large language models: A retrieval augmented generation-based framework. *NPJ Digit Med*. 2024;7(102). doi:10.1038/s41746-024-01091-y

22. Sivarajkumar S, Kelley M, Samolyk-Mazzanti A, Visweswaran S, Wang Y. An empirical evaluation of prompting strategies for large language models in zero-shot clinical natural language processing: Algorithm development and validation study. *JMIR Med Inform*. 12:e55318. doi:10.2196/55318

23. Chen H, Fang Z, Singla Y, Dredze M. Benchmarking large language models on answering and explaining challenging medical questions. Published online 2024. doi:10.48550/arXiv.2402.18060

24. Johnson AEW, Bulgarelli L, Shen L, et al. MIMIC-IV, a freely accessible electronic health record dataset. *Sci Data*. 2023;10(1). doi:10.1038/s41597-022-01899-x

25. Guenette JP, Lynch E, Abbasi N, et al. Recommendations for additional imaging on head and neck imaging examinations: Interradiologist variation and associated factors. *Am J Roentgenol*. 20245;222(5):e2330511. doi:10.2214/AJR.23.30511

26. Derbas LA, Patel KK, Muskula PR, et al. Variability in utilization of diagnostic imaging tests in patients with symptomatic peripheral artery disease. *Int J Cardiol*. 2021;330:200-206. doi:10.1016/j.ijcard.2021.02.014

27. Hughes DR, Jiang M, Duszak Jr. R. A comparison of diagnostic imaging ordering patterns between advanced practice clinicians and primary care physicians following office-based evaluation and management visits. *JAMA Intern Med*. 2015;175(1):101-107. doi:10.1001/jamainternmed.2014.6349

28. The Mosaic Research Team. Introducing DBRX: A new state-of-the-art open LLM. Mosaic Research. March 27, 2024. Accessed July 2, 2024. https://www.databricks.com/blog/introducing-dbrx-new-state-art-open-llm

29. Meta AI. Introducing Meta Llama 3: The most capable openly available LLM to date. Large Language Models. April 18, 2024. Accessed July 2, 2024. https://ai.meta.com/blog/meta-llama-3/

30. Jiang AQ, Sablayrolles A, Roux A, et al. Mixtral of Experts. Published online 2024. doi:10.48550/arXiv.2401.04088

31. CohereForAI/c4ai-command-r-plus. Accessed July 15, 2024. https://huggingface.co/CohereForAI/c4ai-command-r-plus

32. OpenAI, Achiam J, Adler S, et al. GPT-4 technical report. Published online 2024. doi:10.48550/arXiv.2303.08774

33. Anthropic. Claude 3.5 Sonnet. Announcements. June 20, 2024. Accessed July 2, 2024. https://www.anthropic.com/news/claude-3-5-sonnet

34. Park WG, Shaheen NJ, Cohen J, et al. Quality indicators for EGD. *Am J Gastroenterol*. 2015;110(1):60-71. doi:10.1038/ajg.2014.384

35. Patel S, May F, Anderson JC, et al. Updates on age to start and stop colorectal cancer screening: Recommendations from the U.S. Multi-Society Task Force on Colorectal Cancer. *Am J Gastroenterol*. 2022;117(1):57-69. doi:10.14309/ajg.0000000000001548

36. The ASAM clinical practice guideline on alcohol withdrawl management. *J Addict Med*. 2020;14(3):1-72. doi:10.1097/ADM.0000000000000668

37. Xiong G, Jin Q, Lu Z, Zhang A. Benchmarking retrieval-augmented generation for medicine. *Findings Assoc Comp Ling*. Published online 2024:6233-6251.

38. Robertson S, Zaragoza H. The probabilistic relevance framework: BM25 and beyond. *Foundations and Trends in Information Retrieval*. 2009;3(4):333-389. doi:10.1561/1500000019

39. Devlin J, Chang MW, Lee K, Toutanova K. BERT: Pre-training of deep bidirectional transformers for language understanding. *Proc NAACL-HLT*. Published online 2019:4171-4186. doi:10.48550/arXiv.1810.04805

40. Song K, Tan X, Qin T, Lu J, Liu TY. MPNet: Masked and permuted pre-training for language understanding. *Proc Neur Inf Proc Sys*. 2020;(1414):16857-16867. doi:10.5555/3495724.3497138

41. Jin Q, Kim W, Chen Q, et al. MedCPT: Contrastive pre-trained transformers with large-scale PubMed search logs for zero-shot biomedical information retrieval. *Bioinformatics*. 2023;39(11):btad651. doi:10.1093/bioinformatics/btad651

42. Hu EJ, Shen Y, Wallis P, Allen-Zhu Z, Li Y. LoRA: Low-rank adaptation of large language models. *Proc Int Conf Learn Repr*. Published online 2022. https://openreview.net/forum?id=nZeVKeeFYf9

43. Dettmers T, Pagnoni A, Holtzman A, Zettlemoyer L. QLoRA: Efficient finetuning of quantized LLMs. *Proc Neur Inf Proc Sys*. 2024;(441):10088-10115. doi:10.5555/3666122.3666563

44. Virtanen P, Gommers R, Oliphant TE, et al. SciPy 1.0: Fundamental algorithms for scientific computing in Python. *Nat Methods*. 2020;27:261-272. doi:10.1038/s41592-019-0686-2

45. Bergé L. *Efficient Estimation of Maximum Likelihood Models with Multiple Fixed-Effects: The R Package FENmlm*. Department of Economics at the University of Luxembourg; 2018. https://EconPapers.repec.org/RePEc:luc:wpaper:18-13

**SUPPLEMENTARY INFORMATION**

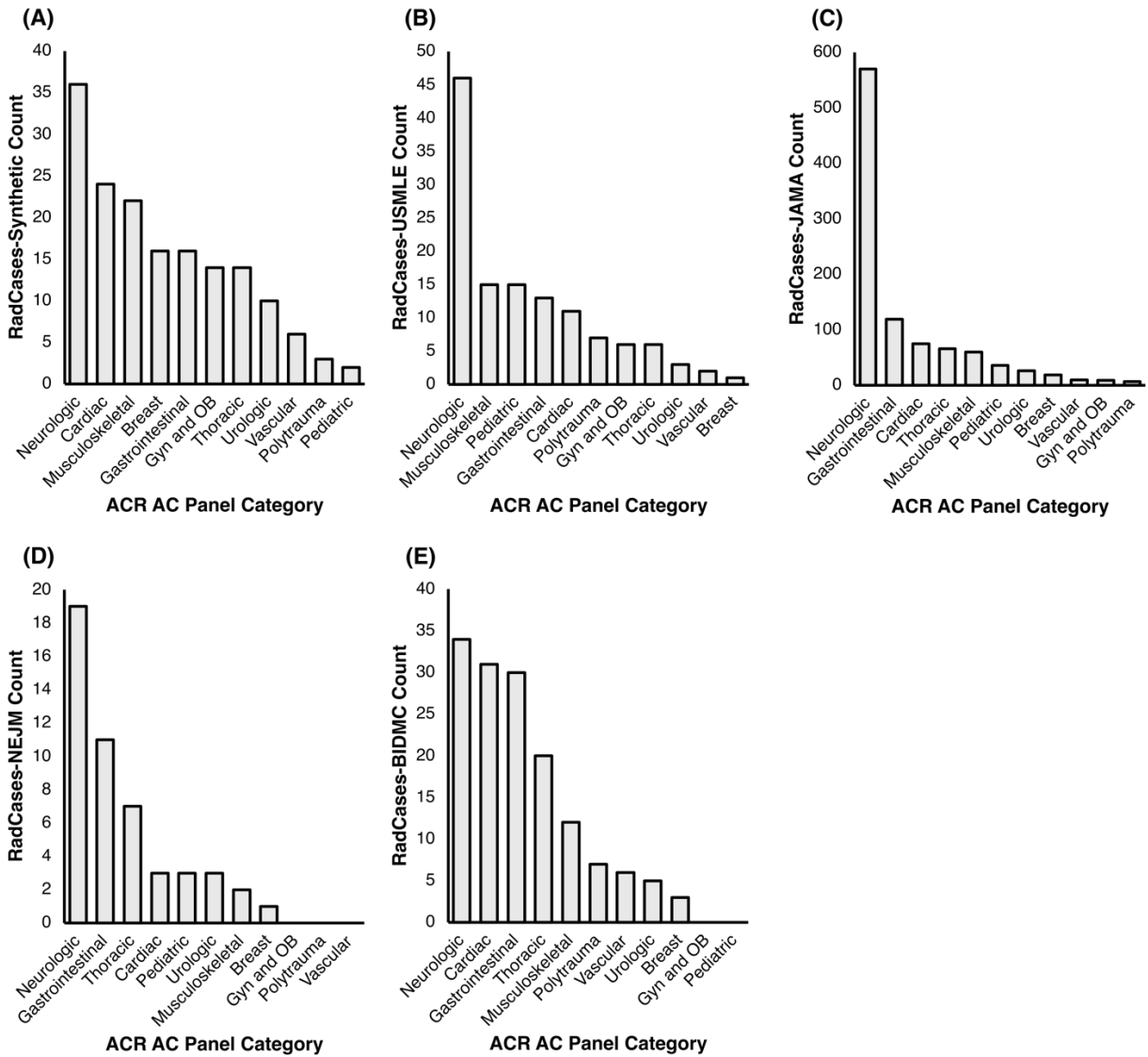

**Supplementary Figure 1: ACR AC Panel Counts in the RadCases Dataset.** To date, there are 224 ACR AC Topics that each have at least one assigned parent ACR AC Panel. Panels are more general categories for conditions, and there are 11 to date: Breast, Cardiac, Gastrointestinal, Gyn and OB, Musculoskeletal, Neurologic, Pediatric, Polytrauma, Thoracic, Urologic, Vascular. To illustrate the distribution of conditions present in the RadCases dataset, we plot the counts of each of these 11 parent ACR AC Panels for the **(A) Synthetic**; **(B) USMLE**; **(C) JAMA**; **(D) NEJM**; and **(E) BIDMC** subsets of the RadCases dataset.

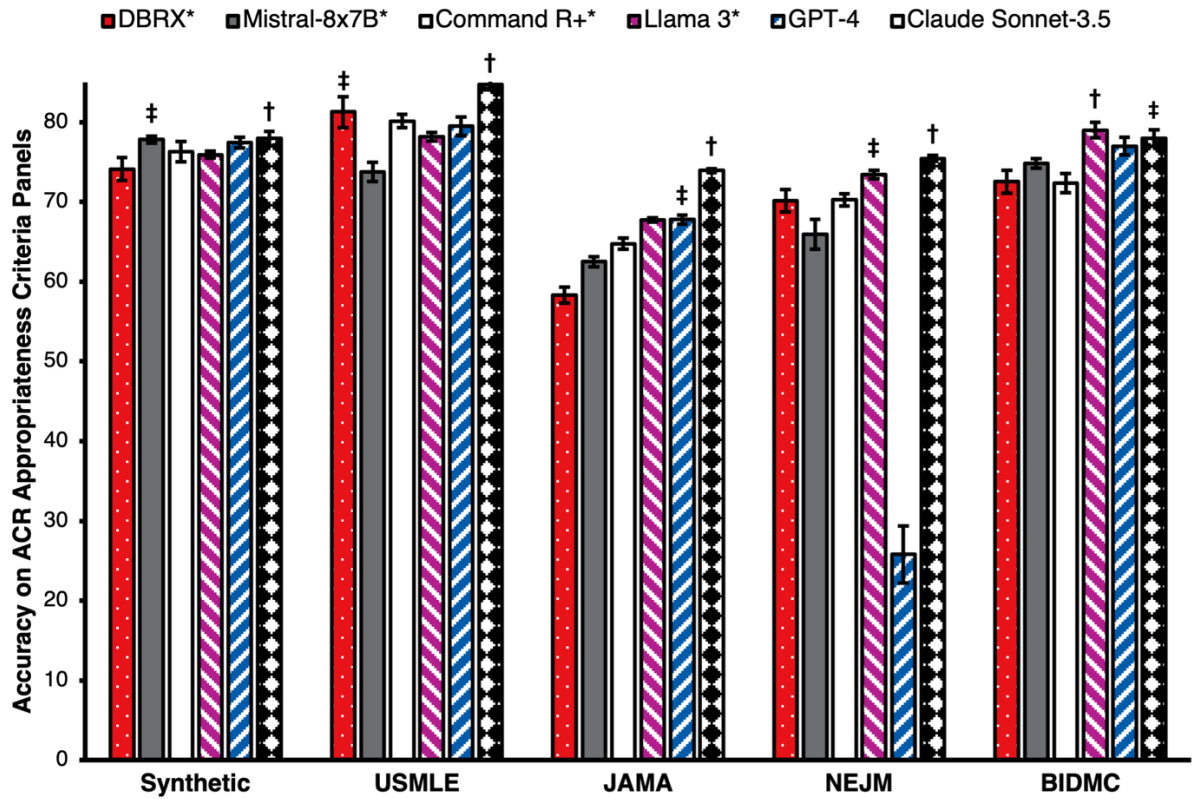

**Supplementary Figure 2: Baseline LLM Performance on ACR AC Panel Classification Using the RadCases Dataset.** In **Figure 1b,** we evaluate six state-of-the-art large language models (LLMs) on their ability to correctly assign 1 of 224 ACR AC Topics to an input one-liner. Here, we include analogous results on the related ACR AC Panel classification task, which queries an LLM to correctly assign 1 of 11 ACR AC Panels to an input one-liner. Because ACR AC Panels are much more coarse-grained when compared to Topics, a language model's accuracy on this task can help assess the model's ability to identify the general body part or organ system affected by pathophysiology. However, accuracy on this task is not helpful for ordering image studies, as there is no clear method for assigning a "correct" imaging study given only an ACR AC Panel. Open-source models are identified by an asterisk, and the best (second best) performing model for a RadCases dataset partition is identified by a dagger (double dagger). Error bars represent ± 95% CI over $n = 5$ independent experimental runs.

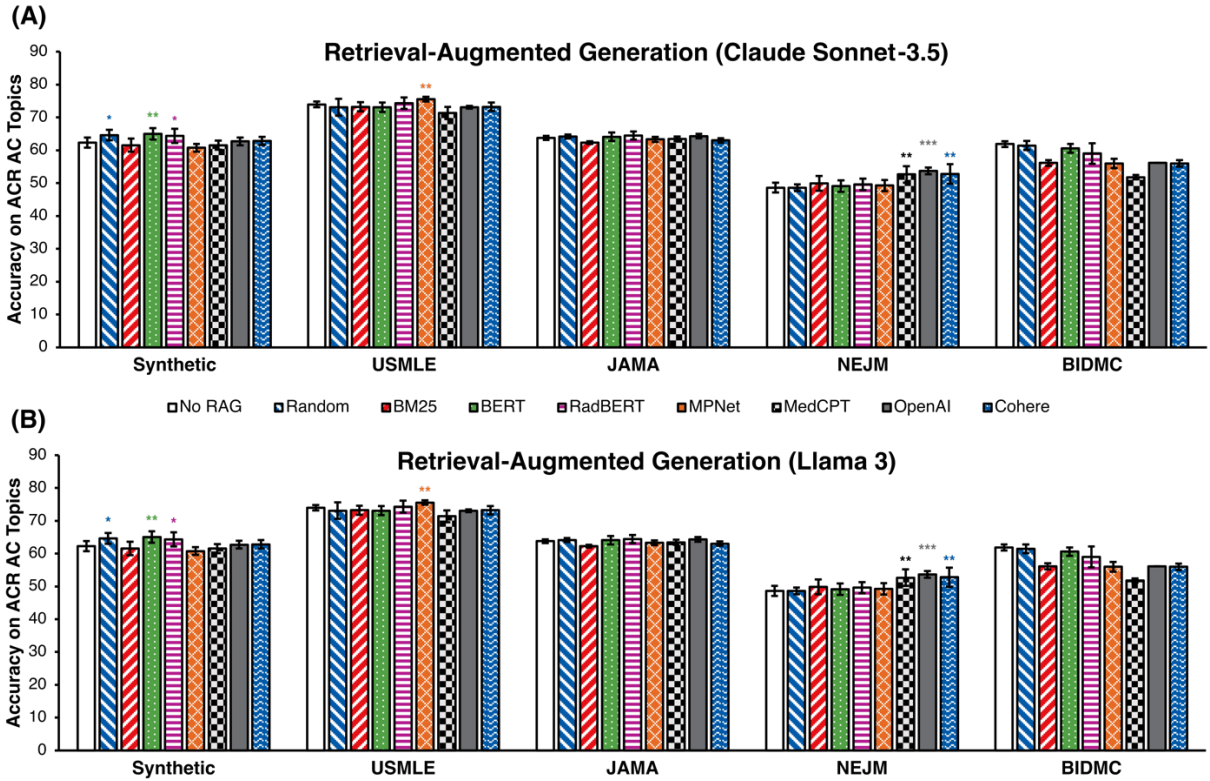

**Supplementary Figure 3: Retrieval-Augmented Generation (RAG) Performance versus Retriever Algorithm.** To optimize RAG for LLM accuracy on the ACR AC Topic classification task, we investigated the use of 8 different retrieval algorithms to use in RAG: (1) **Random**, which randomly documents from the corpus over a uniform probability distribution; (2) Okapi **BM25** bag-of-words retriever; (3) **BERT** and (4) **MPNet** trained on unlabeled, natural language text; (5) **RadBERT** from fine-tuning BERT on radiology text reports; (6) **MedCPT** leveraging a transformer trained on PubMed search logs; and (7) **OpenAI** (text-embedding-3-large) and (8) **Cohere** (cohere.embed-english-v3) embedding models from OpenAI and Cohere for AI, respectively. Using **(A)** Claude Sonnet-3.5 and **(B)** Llama 3, we retrieve $k = 8$ documents from the ACR AC narrative guidelines corpus using each retriever, and compare each method against baseline ACR AC Topic accuracy achieved by each model. Error bars represent ± 95% CI over $n = 5$ independent experimental runs.

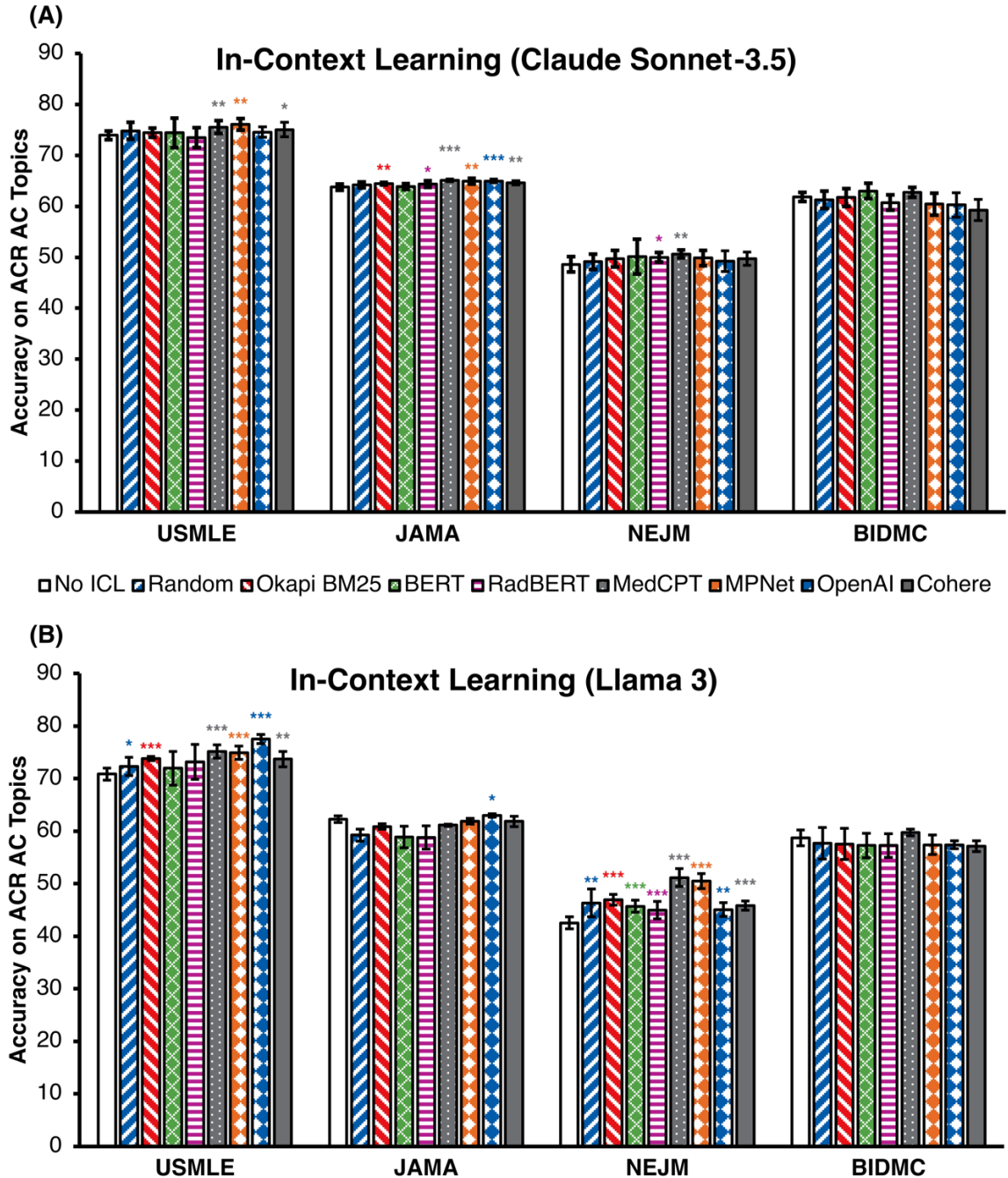

**Supplementary Figure 4: In-Context Learning (ICL) Performance versus Retriever Algorithm.** To optimize ICL for LLM accuracy on the ACR AC Topic classification task, we investigated the use of 8 different retrieval algorithms to use in ICL identical to those explored in RAG (see caption of **Supplementary Fig. 3**). Using **(A)** Claude Sonnet-3.5 and **(B)** Llama 3, we retrieve $k = 4$ example one-liner/Topic pairs from the RadCases-Synthetic dataset corpus using each retriever, and compare each method against baseline ACR AC Topic accuracy achieved by each model. Note that we do not evaluate ICL on the RadCases-Synthetic dataset to avoid data leakage. Error bars represent ± 95% CI over $n = 5$ independent experimental runs.

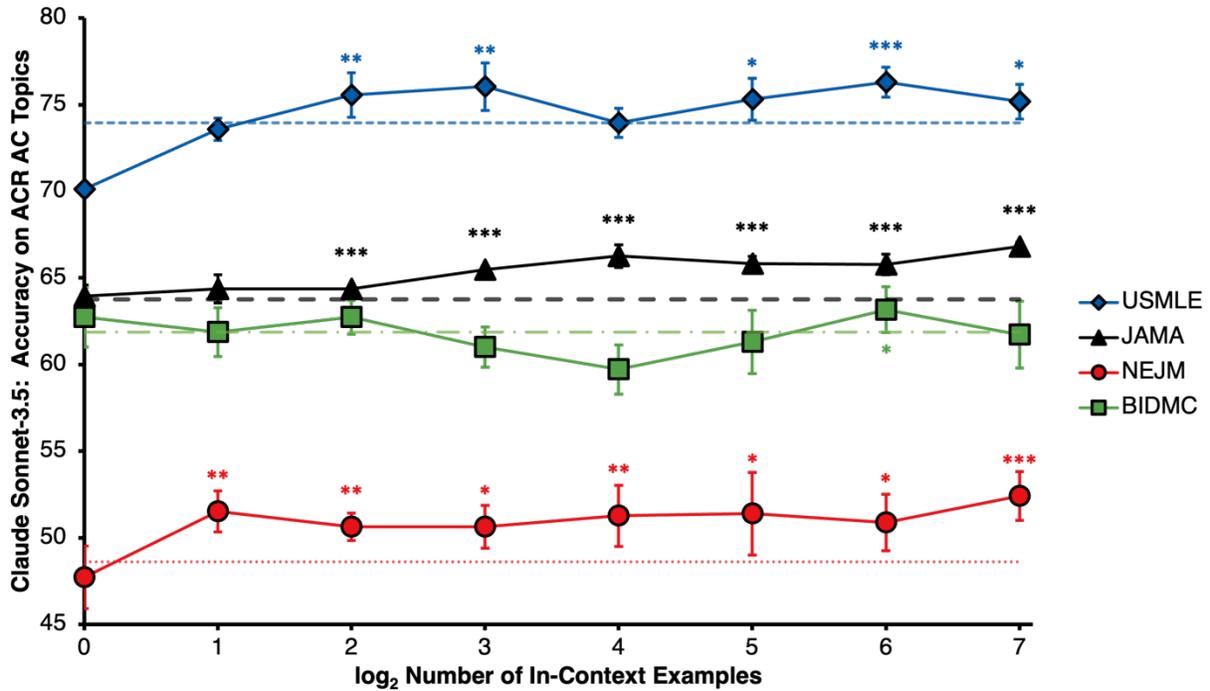

**Supplementary Figure 5: In-Context Learning (ICL) Performance versus Retriever Budget.** Using the subjectively best retriever algorithm evaluated in **Supplementary Fig. 4** (i.e. the MedCPT retriever), we evaluated the effect of changing the number of ICL examples retrieved by the retriever to pass as context to Claude Sonnet-3.5. The blue medium-dashed, black long-dashed, green dotted-dashed, and red dotted horizontal lines correspond to the baseline, no-ICL accuracy scores of Claude Sonnet-3.5 on the USMLE, JAMA, BIDMC, and NEJM subsets of the RadCases dataset, respectively. For the USMLE, JAMA, and NEJM subsets, we find that the performance of the model increases as the number of ICL examples increases from $k = 1$ to $k = 128$. Note that we do not evaluate ICL on the RadCases-Synthetic dataset to avoid data leakage. Error bars represent ± 95% CI over $n = 5$ independent experimental runs.

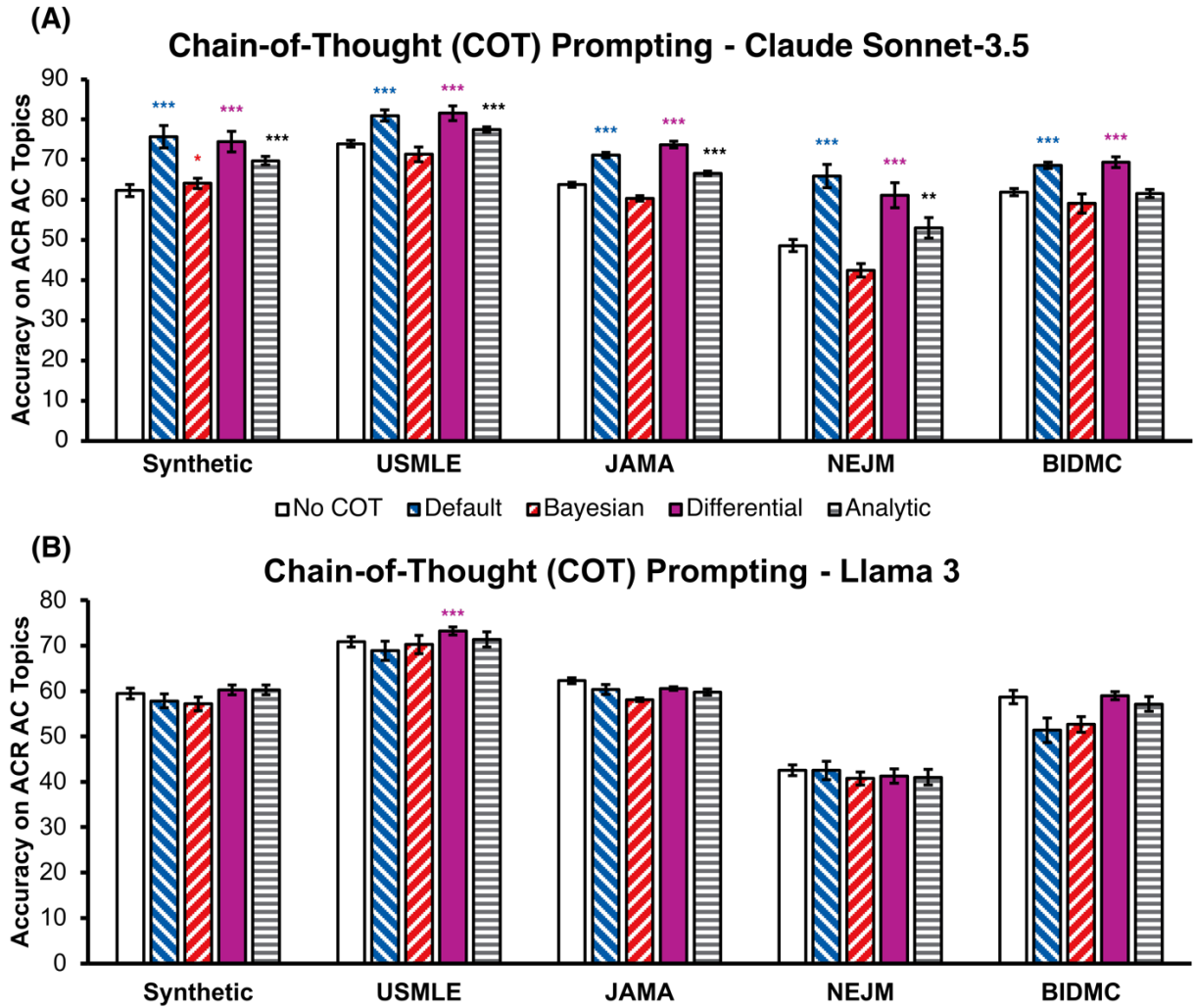

**Supplementary Figure 6: Chain-of-Thought (COT) Prompting Performance versus Reasoning Algorithm.** To optimize COT for LLM accuracy on the ACR AC Topic classification task for both **(A)** Claude Sonnet-3.5 and **(B)** Llama 3, we investigated 4 different COT reasoning methods: (1) **Default** reasoning, which does not specify any particular reasoning strategy for the LLM to use; (2) **Differential** diagnosis reasoning, which encourages the model to reason through a differential diagnosis to arrive at a final prediction; (3) **Bayesian** reasoning, which encourages the model to approximate Bayesian posterior updates over the space of ACR AC Topics based on the clinical patient presentation; and (4) **Analytic** reasoning, which encourages the model to reason through the pathophysiology of the underlying disease process. The exact prompts used in each of these reasoning methods is included in **Supplementary Table 2**. We compare each method against baseline ACR AC Topic accuracy achieved by each model. Error bars represent ± 95% CI over $n = 5$ independent experimental runs.

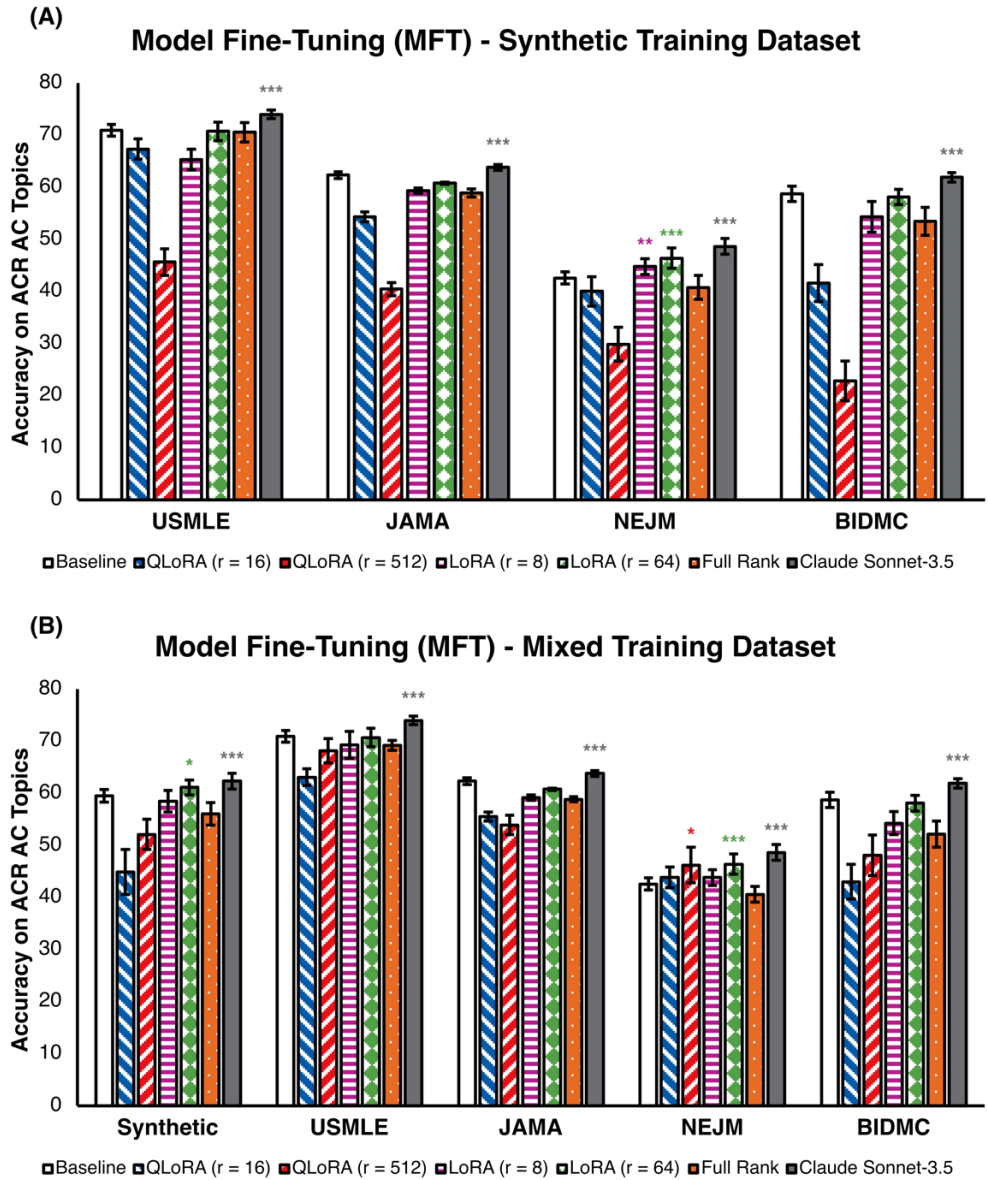

**Supplementary Figure 7: Model Fine-Tuning (MFT) Algorithm Evaluation with Llama 3.** We evaluate 5 different fine-tuning experimental setups in our MFT experiments: quantized low-rank adaptation (**QLoRA**) with a rank of of (1) $r = 16$ and (2) $r = 512$; low-rank adaptation (**LoRA**) with a rank of (3) $r = 8$ and (4) $r = 64$; and (5) **Full Rank** model-finetuning. We use an $\alpha$ scaling value of 8 for all QLoRA and LoRA experiments. To contruct the MFT training dataset, we use either **(A)** all $n = 156$ labelled one-liners from the RadCases-**Synthetic** dataset; or **(B)** a **Mixed** dataset including 50 randomly selected cases from each of the 5 RadCases dataset subsets for a total of $n = 150$ labelled one-liners. The first scenario simulates a setting where we can only fine-tune models on synthetically generated data due to privacy concerns, and the latter scenario simulates a setting where we are able to train on real patient data sampled from the relevant distribution(s) of interest. Note that we do not evaluate MFT on the RadCases-Synthetic dataset in **(A)** to avoid data leakage. Error bars represent ± 95% CI over $n = 5$ independent experimental runs.

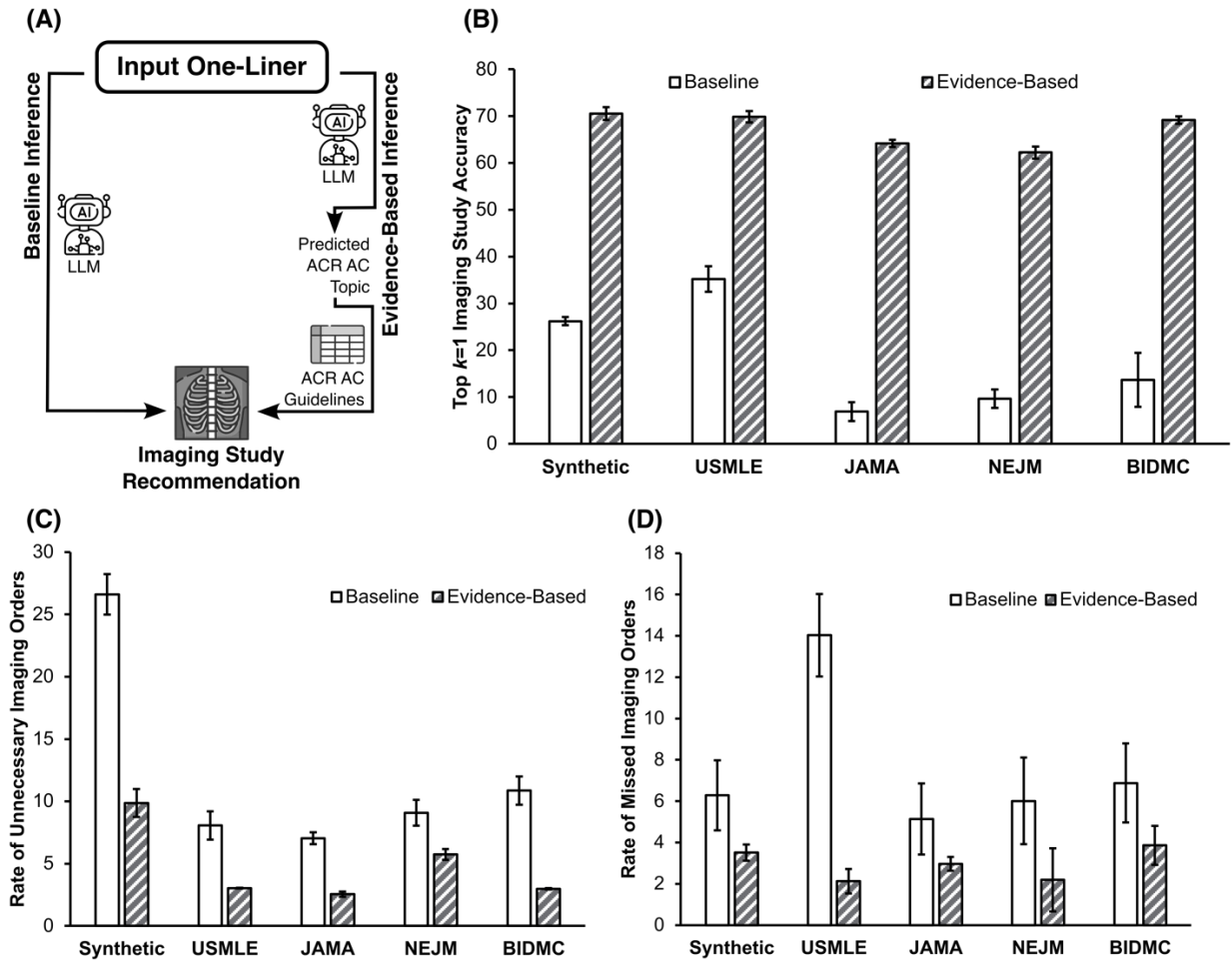

**Supplementary Figure 8: Comparison of Baseline and Evidence-Based Inference Pipelines with Llama 3. (A)** Using our evidence-based inference pipeline identical to that shown in **Fig. 3a** in the main text, we query the Llama 3 to predict the ACR AC Topic most relevant to an input patient one-liner, and programmatically refer to the evidence-based ACR AC guidelines to make the final recommendation for diagnostic imaging (**Evidence-Based**). An alternative approach is the baseline inference pipeline where we query the LLM to recommend a diagnostic imaging study directly without the use of the ACR AC (**Baseline**). Because there was no consistently optimal prompting or fine-tuning strategy that outperformed baseline prompting in **Fig. 2c**, we only empirically evaluated the baseline Evidence-Based inference strategy here. **(B)** Our evidence-based pipeline significantly outperforms the baseline pipeline by up to 57.3% (two-sample, one-tailed, homoscedastic *t*-test; $p < 0.0001$ for all RadCases datasets). At the same time, the also reduce the rates of both **(C)** unnecessary imaging orders and **(D)** missed imaging orders (two-sample, one-tailed, homoscedastic *t*-test; $p < 0.002$ for all RadCases datasets). Error bars represent ± 95% CI over $n = 5$ independent experimental runs.

**Supplementary Figure 9: User Interface for Prospective Study.** The LLM is asked to predict up to three ACR Appropriateness Criteria (AC) Topics that may be relevant for the patient case, and the table of corresponding ACR AC recommendations is displayed as reference to the user. In questions where LLM guidance is not made available, the right column does not show any recommendations and instead displays the message "LLM guidance is not available for this patient scenario."

| **Synthetic Subset ($n_{\text{total}} = 156$)** | **Count (%)** |
|---|---|
| Cardiac > Chest Pain-Possible Acute Coronary Syndrome | 10 (6.41%) |
| Cardiac/Vascular > Suspected Pulmonary Embolism | 8 (5.13%) |
| Gyn and OB > Acute Pelvic Pain in the Reproductive Age Group | 6 (3.85%) |
| Neurologic > Low Back Pain | 6 (3.85%) |
| Breast > Breast Pain | 5 (3.21%) |
| **USMLE Subset ($n_{\text{total}} = 164$)** | **Count (%)** |
| Neurologic > Altered Mental Status, Coma, Delirium, and Psychosis | 17 (10.4%) |
| Neurologic > Headache | 12 (7.32%) |
| Cardiac > Chest Pain-Possible Acute Coronary Syndrome | 9 (5.49%) |
| Polytrauma > Major Blunt Trauma | 9 (5.49%) |
| Cardiac > Dyspnea-Suspected Cardiac Origin | 8 (4.88%) |
| **JAMA Subset ($n_{\text{total}} = 971$)** | **Count (%)** |
| Neurologic > Orbits, Vision, and Visual Loss | 280 (28.8%) |
| Neurologic > Neck Mass/Adenopathy | 48 (4.94%) |
| Neurologic > Staging and Post-Therapy Assessment of Head and Neck Cancer | 42 (4.33%) |
| Neurologic > Headache | 31 (3.19%) |
| Gastrointestinal > Acute Nonlocalized Abdominal Pain | 28 (2.88%) |
| **NEJM Subset ($n_{\text{total}} = 159$)** | **Count (%)** |
| Neurologic > Altered Mental Status, Coma, Delirium, and Psychosis | 11 (6.92%) |
| Neurologic > Orbits, Vision, and Visual Loss | 8 (5.03%) |
| Gastrointestinal > Acute Nonlocalized Abdominal Pain | 7 (4.40%) |
| Neurologic > Headache | 7 (4.40%) |
| Urologic > Renal Failure | 7 (4.40%) |
| **BIDMC Subset ($n_{\text{total}} = 139$)** | **Count (%)** |
| Cardiac > Chest Pain-Possible Acute Coronary Syndrome | 13 (9.35%) |
| Neurologic > Head Trauma | 11 (7.91%) |
| Gastrointestinal > Acute Nonlocalized Abdominal Pain | 10 (7.19%) |
| Neurologic > Altered Mental Status, Coma, Delirium, and Psychosis | 7 (5.04%) |
| Cardiac > Dyspnea-Suspected Cardiac Origin | 6 (4.32%) |

**Supplementary Table 1: Commonly Appearing ACR AC Topics in the RadCases Dataset.** For each of the 5 RadCases dataset subsets, we list the 5 most commonly appearing ACR AC Topics for each of the subsets. The topics are listed as "Panel > Topic," where "Topic" is the ACR AC Topic and "Panel" is the parent ACR AC Panel category(s) of the Topic.

| Gender | Count (%) |
| --- | --- |
| Male | 63 (53.8%) |
| Female | 54 (46.2%) |
| **Age Decade (Years)** | **Count (%)** |
| 10-19 | 2 (1.71%) |
| 20-29 | 5 (4.27%) |
| 30-39 | 7 (5.98%) |
| 40-49 | 15 (12.8%) |
| 50-59 | 27 (23.1%) |
| 60-69 | 30 (25.6%) |
| 70-79 | 27 (23.1%) |
| 80-89 | 4 (3.42%) |
| **Total Number of Patient Cases** | **117** |

**Supplementary Table 2: Simulated Patient Demographics for Retrospective Study Assessing LLMs versus Clinician Performance.** In our retrospective study described in the main text, we analyzed the performance of autonomous LLM agents versus clinicians in ordering diagnostic imaging studies for simulated patient cases crafted from anonymized, de-identified discharge summaries from the MIMIC-IV dataset from Johnson et al.[24] To better simulate actual patient cases, we manually annotated the patient cases to include simulated patient ages and/or genders if they were removed during the original de-identification process. The resulting distributions of these simulated patient variables are shown above.

| Gender | Count (%) |
|---|---|
| Male | 26 (52.0%) |
| Female | 24 (48.0%) |

| Age Decade (Years) | Count (%) |
|---|---|
| 10-19 | 2 (4.00%) |
| 20-29 | 3 (6.00%) |
| 30-39 | 2 (4.00%) |
| 40-49 | 7 (14.0%) |
| 50-59 | 8 (16.0%) |
| 60-69 | 12 (24.0%) |
| 70-79 | 13 (26.0%) |
| 80-89 | 3 (6.0%) |

| **Total Number of Patient Cases** | **50** |
|---|---|

**Supplementary Table 3: Simulated Patient Demographics for Prospective Study Assessing Clinician Performance with versus without LLM-Based Assistance.** In our prospective randomized clinical trial described in the main text, we analyzed the performance of clinicians both with and without LLM-based imaging recommendations in ordering diagnostic imaging studies for simulated patient one-liners. These one-liners were crafted from anonymized, de-identified discharge summaries from the MIMIC-IV dataset from Johnson et al.[24] To better simulate actual patient cases, we manually re-introduced include simulated patient ages and/or genders if they were removed during the original de-identification process. The resulting distributions of these simulated patient variables are shown above.

|  | (1) | (2) |
|---|---|---|
| **Gender** | Count (%) | Count (%) |
| Male | 4 (25.0) | 9 (64.3) |
| Female | 12 (75.0) | 5 (35.7) |
| **Stage of Medical Training** | Count (%) | Count (%) |
| Third- or Fourth- Year U.S. Medical Student | 14 (87.5) | 9 (64.3) |
| U.S. Emergency Medicine Resident Physician | 2 (12.5) | 5 (35.7) |
| **Prior Experience Using AI in Everyday Life** | Count (%) | Count (%) |
| No Experience or A Little Experience | 8 (50.0) | 9 (64.3) |
| Some Experience or A Lot of Experience | 8 (50.0) | 5 (35.7) |
| **Overall Sentiment of the Use of AI in Healthcare** | Count (%) | Count (%) |
| Negative | 2 (12.5) | 2 (14.3) |
| Neutral or Positive | 14 (87.5) | 12 (85.7) |

**Supplementary Table 4: Study Participant Demographic Information in Prospective Study Assessing Clinician Performance with versus without LLM-Based Assistance.** Demographic and self-reported pre-study questionnaire information of the clinician study participants in our prospective study detailed in the main text are summarized above. Column (1) describes the participants randomized to the Timed study arm described in **Appendix A**, and column (2) describes the participants randomized to the Untimed study arm in **Appendix A**.

|  | (1) | (2) | (3) | (4) |
|---|---|---|---|---|
| LLM Guidance Available | 0.081* (0.028) | 0.081* (0.028) | -0.089* (0.037) | -0.089* (0.037) |
| p Value | 0.011 | 0.011 | 0.032 | 0.032 |
| R2 | 0.138 | 0.138 | 0.121 | 0.121 |
| Number of Observations | 800 | 800 | 700 | 700 |

**Supplementary Table 5: Accuracy Scores of Clinicians with and without LLM-Generated Recommendations.** The treatment effect of offering LLM-generated diagnostic imaging recommendations is analyzed according to **Equation 1**. The accuracy score is a binary dependent variable equal to 1 if the clinician orders a ground-truth imaging study according to the ACR Appropriateness Criteria, and 0 otherwise. The regression coefficients are shown as mean (standard error). Columns (1) and (2) correspond to the timed experimental arm where participants are required to answer questions at an average rate no slower than 1 question per minute, while columns (3) and (4) correspond to the separate untimed experimental arm where participants can answer questions at their own pace in one sitting. Odd- (even-) numbered columns do not (do) factor in the fixed effects participant self-reported personal experience with AI and personal sentiment on the use of AI in medicine, which are both modeled as binary variables, into the regression model. *Denotes $p < 0.05$.

|  | (1) | (2) |
|---|---|---|
| LLM Guidance Available | 0.141** (0.043) | 0.107** (0.031) |
| p Value | 0.005 | 0.004 |
| Prior Experience Using AI | 0.043 (0.051) | -0.169** (0.055) |
| p Value | 0.407 | 0.009 |
| Positive Sentiment About AI | -0.219** (0.063) | -0.086* (0.031) |
| p Value | 0.004 | 0.016 |
| R2 | 0.380 | 0.365 |
| Observations | 800 | 700 |

**Supplementary Table 6: LLM Agreement Scores of Clinicians with and without LLM-Generated Recommendations.** The treatment effect of offering LLM-generated diagnostic imaging recommendations is analyzed according to

$$z_{s,q} = \beta_0 + (\beta_1 * \text{WithLLMGuidance}_{s,q}) + (\beta_2 * \text{PriorExperienceUsingAI}_s) + (\beta_3 * \text{PositiveSentimentAboutAI}_s) + \theta_q + \chi_s + \varepsilon_{s,q}$$

where $z_{s,q}$ is the agreement score represented as a binary dependent variable equal to 1 if the clinician and LLM recommend the same imaging study and 0 otherwise; $\theta_q$ is the fixed effects of study question $q$; $\chi_s$ the fixed effects of study participant $s$; and $\varepsilon_{s,q}$ is the error term. All the independent variables in this regression model are binarized take on values in $\{0, 1\}$. The regression coefficients are shown as mean (standard error). Column (1) corresponds to the timed experimental arm where participants are required to answer questions at an average rate no slower than 1 question per minute, while column (2) corresponds to the separate untimed experimental arm where participants can answer questions at their own pace in one sitting. *Denotes $p < 0.05$. **Denotes $p < 0.01$.

|  | (1) | (2) | (3) | (4) |
|---|---|---|---|---|
| LLM Guidance Available | 0.008 (0.009) | 0.008 (0.009) | -0.005 (0.011) | -0.005 (0.011) |
| p Value | 0.412 | 0.418 | 0.633 | 0.630 |
| R2 | 0.057 | 0.054 | 0.061 | 0.059 |
| Number of Observations | 800 | 800 | 700 | 750 |

**Supplementary Table 7: False Positive Rates of Clinicians with and without LLM-Generated Recommendations.** The treatment effect of offering LLM-generated diagnostic imaging recommendations is analyzed according to **Equation 1**. A false positive is a binary dependent variable equal to 1 if the clinician orders an unnecessary imaging study according to the ACR Appropriateness Criteria, and 0 otherwise. The regression coefficients are shown as mean (standard error). Columns (1) and (2) correspond to the timed experimental arm where participants are required to answer questions at an average rate no slower than 1 question per minute, while columns (3) and (4) correspond to the separate untimed experimental arm where participants can answer questions at their own pace in one sitting. Odd- (even-) numbered columns do not (do) factor in the fixed effects of participant self-reported personal experience with AI and personal sentiment on the use of AI in medicine, which are both modeled as binary variables, into the regression model.

|  | (1) | (2) | (3) | (4) |
|---|---|---|---|---|
| LLM Guidance Available | -0.019 (0.023) | -0.019 (0.023) | 0.001 (0.021) | 0.001 (0.021) |
| p Value | 0.440 | 0.431 | 0.952 | 0.951 |
| R2 | 0.221 | 0.180 | 0.227 | 0.218 |
| Number of Observations | 800 | 800 | 700 | 700 |

**Supplementary Table 8: False Negative Rates of Clinicians with and without LLM-Generated Recommendations.** The treatment effect of offering LLM-generated diagnostic imaging recommendations is analyzed according to **Equation 1**. A false negative is a binary dependent variable equal to 1 if the clinician orders no imaging study even when diagnostic imaging is warranted according to the ACR Appropriateness Criteria, and 0 otherwise. The regression coefficients are shown as mean (standard error). Columns (1) and (2) correspond to the timed experimental arm where participants are required to answer questions at an average rate no slower than 1 question per minute, while columns (3) and (4) correspond to the separate untimed experimental arm where participants can answer questions at their own pace in one sitting. Odd- (even-) numbered columns do not (do) factor in the fixed effects of participant self-reported personal experience with AI and personal sentiment on the use of AI in medicine, which are both modeled as binary variables, into the regression model.

| (Q1) In your *personal life*, how much prior experience do you have with using machine learning models, such as ChatGPT or other AI tools? | Count (%) |
|---|---|
| No prior experience | 2 (7.4) |
| A little prior experience | 17 (63.0) |
| Some prior experience | 8 (29.6) |
| A lot of prior experience | 0 (0.0) |
| **(Q2) In your *personal role*, how much prior experience do you have with using machine learning models, such as ChatGPT or other AI tools?** | **Count (%)** |
| No prior experience | 23 (85.2) |
| A little prior experience | 4 (14.8) |
| Some prior experience | 0 (0.0) |
| A lot of prior experience | 0 (0.0) |
| **(Q3) AI can help improve patient care and clinical workflows in the future.** | **Count (%)** |
| I strongly disagree | 0 (0.0) |
| I somewhat disagree | 0 (0.0) |
| I am neutral | 1 (3.7) |
| I somewhat agree | 15 (55.6) |
| I strongly agree | 11 (40.7) |
| **(Q4) I am scared about the potential unknown impact of AI on healthcare.** | **Count (%)** |
| I strongly disagree | 1 (3.7) |
| I somewhat disagree | 8 (29.6) |
| I am neutral | 2 (7.4) |
| I somewhat agree | 13 (48.2) |
| I strongly agree | 3 (11.1) |
| **(Q5) Overall, how positive or negative do you feel about the potential use of AI in medicine?** | **Count (%)** |
| Very negative | 0 (0.0) |
| Somewhat negative | 3 (11.1) |
| Neutral | 2 (7.4) |
| Somewhat positive | 18 (66.7) |
| Very positive | 4 (14.8) |

**Supplementary Table 9: Study Participant Pre-Study Survey Results.** All study participants were asked to complete an anonymized survey of 5 multiple-choice questions prior to beginning the study. The results of (Q1) and (Q5) were used to define the $\text{PriorExperienceUsingAI}_s$ and $\text{PositiveSentimentAboutAI}_s$ binary variables used in the regression models, respectively. For a subject $s$, $\text{PriorExperienceUsingAI}_s$ is equal to 1 if the subject answers "Some experience" or "A lot of prior experience" to (Q1) and 0 otherwise. Similarly, $\text{PositiveSentimentAboutAI}_s$ is equal to 1 if the subject answers "Neutral", "Somewhat positive", or "Very positive" to (Q5) and 0 otherwise.

**APPENDIX A: ADDITIONAL RESULTS FOR PROSPECTIVE CLINICIAN-AI STUDY**

In this section, we offer additional experimental results for our prospective study with U.S. medical students and emergency medicine resident physicians. The main text of our work describes the results of our **Timed** experimental arm, where participants were required to complete the study at an average rate of no slower than 1 question per minute. We also ran a separate, **Untimed** experimental arm, where participants were no constraints were imposed on the rate of completion so long as the study was completed in one sitting. Participants were randomized in exactly one of the two experimental arms: of the 30 participants who completed the study, 16 (14) were assigned to the Timed (Untimed) arm. On average, participants in the Timed experimental arm completed the study in 36.74 minutes (95% CI: [29.88 – 43.60]), while participants in the Untimed experimental arm completed the study in 46.80 minutes (95% CI: [33.90 – 59.70]). The overall accuracy of the Timed arm participants was 20.4% (95% CI: [17.6% – 23.2%]), the accuracy of the Untimed arm participants was 20.9% (95% CI: [17.8% – 23.9%]). In contrast, the accuracy of the optimized language model along was 72.8% on the prospective study task, highlighting the difficulty of this task. Knowledge of the performance of language model on the study tasks was not made available to the study participants.

In **Supplementary Table 5** and in the main text, we describe a statistically significant improvement in the accuracy of ordered diagnostic imaging studies by study participants when LLM-generated guidance is offered in the Timed experimental arm. Interestingly, we found that the *inverse* is true in the Untimed arm: participant accuracy *decreased* with statistical significance with LLM-generated guidance was available ($\beta_1$ = -0.089; 95% CI: [-0.170 – -0.009]; $p$ = 0.032). We hypothesize that this may be because participants have more time to carefully think through cases and consult external resources in the Untimed arm. The absence of the "pressure" imposed by a time limit may also have psychological impacts in clinical decision making that are outside the scope of this work. Regardless, ostensibly paradoxical experimental findings—such as the results in the Untimed study arm—have been previously reported in related work studying human-computer interaction with generative AI systems; for example, Dell'Acqua (2022)[S1] and Dell'Acqua et al. (2023)[S2] describe how AI systems can adversely impact expert performance on specialized tasks under certain conditions, and Bastani et al. (2024)[S3] characterize how generative AI tools deter student learning if key guardrails are not properly implemented. We believe these results emphasize the importance of carefully studying how different clinical workflows are uniquely impacted generative AI tools in future work.

**Supplementary Tables 6-8** describe the effect of LLM-generated recommendations on (1) LLM agreement; (2) false positive rate; and (3) false negative rate. For both the Timed and Untimed experimental arms, we observe that LLM-generated recommendations increase LLM agreement and do not affect the false positive or false negative rates of image ordering. Interestingly, a self-reported positive sentiment regarding AI in medicine was associated with *lower* LLM agreement scores in both the Timed ($\beta_3$ = -0.219; 95% CI: [-0.353 – -0.084]; $p$ = 0.004) and Untimed ($\beta_3$ = -0.086; 95% CI: [-0.153 – -0.019]; $p$ = 0.016) experimental arms.

[S1] Dell'Acqua F. Falling asleep at the wheel: Human/AI collaboration in a field experiment on HR recruiters. Working paper. (2022). URL
[S2] Dell'Acqua F, McFowland E, Mollick ER, et al. Navigating the jagged technological frontier: Field experimental evidence of the effects of AI on knowledge worker productivity and quality. Harvard Business School Technology & Operations Mgt. Unit Working Paper 24-013. (2023). doi: 10.2139/ssrn.4573321
[S3] Bastani H, Bastani O, Sungu A, et al. Generative AI can harm learning. The Wharton School Research Paper. (2024). doi: 10.2139/ssrn.4895486

**APPENDIX B: LARGE LANGUAGE MODEL SYSTEM AND USER PROMPTS**

For the prompts listed below, the following formatting string rules apply:
- <l categories l> is equal to the list of all the ACR AC Panels concatenated using the "; " string for the ACR AC Panel classification task; the list of all the ACR AC Topics concatenated using the "; "

string for the ACR AC Topic classification task; and the list of all the Imaging Studies present in the ACR AC concatenated using the "; " string for the Imaging Study classification task.
- <| example |> is equal to "Thoracic" for the ACR AC Panel classification task; "Lung Cancer Screening" for the ACR AC Topic classification task; and "CT chest without IV contrast screening" for the Imaging Study classification task.
- <| case |> is equal to the patient one-liner case description.
- <| context |> is equal to the retrieved corpus documents retrieved by the retrieval algorithm used in either RAG or ICL prompting. The documents are separated by two new-line characters to form the context string.

**System Prompt for Baseline Prompting, In-Context Learning (ICL) Prompting, and Model Fine-Tuning (MFT) for ACR AC Panel and ACR AC Topic Classification Tasks**

You are a clinical decision support tool that classifies patient one-liners into categories. Classify each query into one of the following categories. Provide your output in JSON format with the single key "answer"

Categories: <| categories |>

Example: 49M with HTN, IDDM, HLD, and 20 pack-year smoking hx p/w 4 mo hx SOB and non-productive cough.
Answer: {"answer": "<| example |>"}

**System Prompt for Baseline Prompting, In-Context Learning (ICL) Prompting, and Model Fine-Tuning (MFT) for Imaging Study Classification Task**

You are a clinical decision support tool that determines the best imaging studies to order for patients. Choose the best imaging study the following options. If no imaging is required, return "None". Provide your output in JSON format with the single key "answer"

Categories: <| categories |>

Example: 49M with HTN, IDDM, HLD, and 20 pack-year smoking hx p/w 4 mo hx SOB and non-productive cough.
Answer: {"answer": "<| example |>"}

**System Prompt for Retrieval-Augmented Generation (RAG) Prompting for ACR AC Panel, ACR AC Topic, and Imaging Study Classification Tasks**

You are a clinical decision support tool that classifies patient one-liners into categories. Classify each query into one of the following categories. You will be given context that might be helpful, but you can ignore the context if it is not helpful. Provide your output in JSON format with the single key "answer"

Categories: <| categories |>

Example: 49M with HTN, IDDM, HLD, and 20 pack-year smoking hx p/w 4 mo hx SOB and non-productive cough.
Answer: {"answer": "<| example |>"}

**System Prompt for Chain-of-Thought (COT) Prompting Using Default Reasoning for ACR AC Panel, ACR AC Topic, and Imaging Study Classification Tasks**

You are a clinical decision support tool that classifies patient one-liners into categories. Classify each query into one of the following categories. Provide your output in JSON format {"answer": [YOUR CLASSIFICATION], "rationale": [YOUR STEP-BY-STEP REASONING]}

Categories: <| categories |>

Use step-by-step deduction to identify the correct classification.

**System Prompt for Chain-of-Thought (COT) Prompting Using Bayesian Reasoning for ACR AC Panel, ACR AC Topic, and Imaging Study Classification Tasks**

You are a clinical decision support tool that classifies patient one-liners into categories. Classify each query into one of the following categories. Provide your output in JSON format {"answer": [YOUR CLASSIFICATION], "rationale": [YOUR STEP-BY-STEP REASONING]}

Categories: <| categories |>

In your rationale, use step-by-step Bayesian Inference to create a prior probability that is updated with new information in the history to produce a posterior probability and determine the final classification.

**System Prompt for Chain-of-Thought (COT) Prompting Using Differential Diagnosis Reasoning for ACR AC Panel, ACR AC Topic, and Imaging Study Classification Tasks**

You are a clinical decision support tool that classifies patient one-liners into categories. Classify each query into one of the following categories. Provide your output in JSON format {"answer": [YOUR CLASSIFICATION], "rationale": [YOUR STEP-BY-STEP REASONING]}

Categories: <| categories |>

Use step-by-step deduction to first create a differential diagnosis, and select the answer that is most consistent with your reasoning.

**System Prompt for Chain-of-Thought (COT) Prompting Using Analytic Reasoning for ACR AC Panel, ACR AC Topic, and Imaging Study Classification Tasks**

You are a clinical decision support tool that classifies patient one-liners into categories. Classify each query into one of the following categories. Provide your output in JSON format {"answer": [YOUR CLASSIFICATION], "rationale": [YOUR STEP-BY-STEP REASONING]}

Categories: <| categories |>

Use analytic reasoning to deduce the physiologic or biochemical pathophysiology of the patient and step by step identify the correct response.

**User Prompt for Baseline Prompting, Chain-of-Thought (COT) Prompting, and Model Fine-Tuning (MFT) for ACR AC Panel and ACR AC Topic Classification Tasks If Requesting 1 LLM Prediction**

Patient Case: <| case |>

Which category best describes the patient's chief complaint?

**User Prompt for Baseline Prompting, Chain-of-Thought (COT) Prompting, and Model Fine-Tuning (MFT) for ACR AC Panel and ACR AC Topic Classification Tasks If Requesting 3 LLM Predictions**

Patient Case: <| case |>

Select up to 3 categories that best describe the patient's chief complaint.

**User Prompt for Baseline Prompting, Chain-of-Thought (COT) Prompting, and Model Fine-Tuning (MFT) for Imaging Study Classification Tasks If Requesting 1 LLM Prediction**

Patient Case: <| case |>

Which imaging study (if any) is most appropriate for this patient?

**User Prompt for Baseline Prompting, Chain-of-Thought (COT) Prompting, and Model Fine-Tuning (MFT) for Imaging Study Classification Tasks If Requesting 3 LLM Predictions**

Patient Case: <| case |>

Select up to 3 imaging studies that are most appropriate for this patient.

**User Prompt for Retrieval-Augmented Generation (RAG) Prompting for ACR AC Panel and ACR AC Topic Classification Tasks If Requesting 1 LLM Prediction**

Here is some context for you to consider:
<| context |>

### User:
Patient Case: <| case |>

Which category best describes the patient's chief complaint?

### Assistant:

**User Prompt for Retrieval-Augmented Generation (RAG) Prompting for ACR AC Panel and ACR AC Topic Classification Tasks If Requesting 3 LLM Predictions**

Here is some context for you to consider:
<| context |>

### User:
Patient Case: <| case |>

Select up to 3 categories that best describe the patient's chief complaint.

### Assistant:

**User Prompt for Retrieval-Augmented Generation (RAG) Prompting for Imaging Study Classification Tasks If Requesting 1 LLM Prediction**

Here is some context for you to consider:
<| context |>

### User:
Patient Case: <| case |>

Which imaging study (if any) is most appropriate for this patient?

### Assistant:

**User Prompt for Retrieval-Augmented Generation (RAG) Prompting for Imaging Study Classification Tasks If Requesting 3 LLM Predictions**

Here is some context for you to consider:
<| context |>

### User:
Patient Case: <| case |>

Select up to 3 imaging studies that are most appropriate for this patient.

### Assistant:

**User Prompt for In-Context Learning (ICL) Prompting for ACR AC Panel and ACR AC Topic Classification Tasks If Requesting 1 LLM Prediction**

<| context |>

### User:
Patient Case: <| case |>

Which category best describes the patient's chief complaint?

### Assistant:

**User Prompt for In-Context Learning (ICL) Prompting for ACR AC Panel and ACR AC Topic Classification Tasks If Requesting 3 LLM Predictions**

<| context |>

### User:
Patient Case: <| case |>

Select up to 3 categories that best describe the patient's chief complaint.

### Assistant:

**User Prompt for In-Context Learning (ICL) Prompting for Imaging Study Classification Tasks If Requesting 1 LLM Prediction**

<| context |>

### User:
Patient Case: <| case |>

Which imaging study (if any) is most appropriate for this patient?

### Assistant:

**User Prompt for In-Context Learning (ICL) Prompting Prompting for Imaging Study Classification Tasks If Requesting 3 LLM Predictions**

<| context |>

### User:
Patient Case: <| case |>

Select up to 3 imaging studies that are most appropriate for this patient.

### Assistant:

**APPENDIX C: EXAMPLE CURATION OF LONG-FORM PATIENT SCENARIOS FROM DISCHARGE SUMMARIES FOR RETROSPECTIVE COMPARISON AGAINST CLINICIANS AND PROSPECTIVE CLINICIAN-AI STUDIES**

In our retrospective study described in the main text, we analyzed the performance of autonomous LLM agents versus clinicians in ordering diagnostic imaging studies for simulated patient cases. These cases were longer than traditional "one-liners" to best approximate the amount of information available to the corresponding clinicians when they ordered imaging studies for patients during the emergency room encounter. As a result, the process to manually convert discharge summaries from the MIMIC-IV dataset from Johnson et al[24] to input cases for LLM evaluation was performed through human annotation. We show two examples below of the annotation process to help illustrate how these cases were crafted. Text that is highlighted or written in red were removed from the original discharge summary, and text that is highlighted in green were added to the discharge summary to form the final patient case.

**LLM Input Example 1**

Name: ___        Unit No: ___

Admission Date: ___        Discharge Date: ___

Date of Birth: ___        Sex: F

Service: MEDICINE

Allergies:
No Known Allergies / Adverse Drug Reactions

Attending: ___

Chief Complaint:
Worsening ABD distension and pain

Major Surgical or Invasive Procedure:
Paracentesis

History of Present Illness:
___ 47F w/ HCV cirrhosis c/b ascites, hiv on ART, h/o IVDU, COPD, bioplar, PTSD, presented from OSH ED with worsening abd distension over past week. Pt reports self-discontinuing lasix and spirnolactone

___ 3 weeks ago, because she feels like "they don't do anything" and that she "doesn't want to put more chemicals in her." She does not follow Na-restricted diets. In the past week, she notes that she has been having worsening abd distension and discomfort. She denies ___ edema, or SOB, or orthopnea. She denies f/c/n/v, d/c, dysuria. She had food poisoning a week ago from eating stale cake (n/v 20 min after food ingestion), which resolved the same day. She denies other recent illness or sick contacts. She notes that she has been noticing gum bleeding while brushing her teeth in recent weeks. she denies easy bruising, melena, BRBPR, hemetesis, hemoptysis, or hematuria.  Because of her abd pain, she went to OSH ED and was transferred to ___ HUP for further care. Per ED report, pt has brief period of confusion - she did not recall the ultrasound or bloodwork at
osh. She denies recent drug use or alcohol use. She denies feeling confused, but reports that she is forgetful at times.  In the ED, initial vitals were 98.4 70 106/63 16 97%RA
Labs notable for ALT/AST/AP ___ ___: ___, Tbili1.6, WBC 5K, platelet 77, INR 1.6

Past Medical History:
1. HCV Cirrhosis
2. No history of abnormal Pap smears.
3. She had calcification in her breast, which was removed
previously and per patient not, it was benign.
4. For HIV disease, she is being followed by Dr. ___ Dr. ___.
5. COPD
6. Past history of smoking.
7. She also had a skin lesion, which was biopsied and showed  skin cancer per patient report and is scheduled for a complete  removal of the skin lesion in ___ of this year.
8. She also had another lesion in her forehead with purple  discoloration. It was biopsied to exclude the possibility of  ___'s sarcoma, the results is pending.
9. A 15 mm hypoechoic lesion on her ultrasound on ___  and is being monitored by an MRI.
10. History of dysplasia of anus in ___.
11. Bipolar affective disorder, currently manic, mild, and PTSD.
 12. History of cocaine and heroin use.

Social History:
___
Family History:
She a total of five siblings, but she is not  talking to most of them. She only has one brother that she is in touch with and lives in ___. She is not aware of any  known GI or liver disease in her family.  Her last alcohol consumption was one drink two months ago. No  regular alcohol consumption. Last drug use ___ years ago. She  quit smoking a couple of years ago.

Physical Exam:
VS: 98.1 107/61 78 18 97RA
General: in NAD
HEENT: CTAB, anicteric sclera, OP clear
Neck: supple, no LAD
CV: RRR,S1S2, no m/r/g
Lungs: CTAb, prolonged expiratory phase, no w/r/r
Abdomen: distended, mild diffuse tenderness, +flank dullness, cannot percuss liver/spleen edge ___ distension
GU: no foley
Ext: wwp, no c/e/e, + clubbing
Neuro: AAO3, converse normally, able to recall 3 times after 5 minutes, CN II-XII intact

Discharge:

PHYSICAL EXAMINATION:
VS: 98 105/70 95
General: in NAD
HEENT: anicteric sclera, OP clear
Neck: supple, no LAD
CV: RRR,S1S2, no m/r/g
Lungs: CTAb, prolonged expiratory phase, no w/r/r
Abdomen: distended but improved, TTP in RUQ,
GU: no foley
Ext: wwp, no c/e/e, + clubbing
Neuro: AAO3,  CN II-XII intact

Pertinent Results:
___ 10:25PM   GLUCOSE-109* UREA N-25* CREAT-0.3* SODIUM-138
POTASSIUM-3.4 CHLORIDE-105 TOTAL CO2-27 ANION GAP-9
___ 10:25PM   estGFR-Using this
___ 10:25PM   ALT(SGPT)-100* AST(SGOT)-114* ALK PHOS-114*
TOT BILI-1.6*
___ 10:25PM   LIPASE-77*
___ 10:25PM   ALBUMIN-3.3*
___ 10:25PM   WBC-5.0# RBC-4.29 HGB-14.3 HCT-42.6 MCV-99*
MCH-33.3* MCHC-33.5 RDW-15.7*
___ 10:25PM   NEUTS-70.3* LYMPHS-16.5* MONOS-8.1 EOS-4.2*
BASOS-0.8
___ 10:25PM   PLT COUNT-71*
___ 10:25PM   ___ PTT-30.9 ___
___ 10:25PM   ___
.

CXR: No acute cardiopulmonary process.
U/S:
1. Nodular appearance of the liver compatible with cirrhosis.
Signs of portal  hypertension including small amount of ascites and splenomegaly.

2. Cholelithiasis.
3. Patent portal veins with normal hepatopetal flow.  Diagnostic para attempted in the ED, unsuccessful.
On the floor, pt c/o abd distension and discomfort.

Brief Hospital Course:
___ HCV cirrhosis c/b ascites, hiv on ART, h/o IVDU, COPD, bioplar, PTSD, presented from OSH ED
with worsening abd distension over past week and confusion.

# Ascites - p/w worsening abd distension and discomfort for last week. likely ___ portal HTN given
underlying liver disease, though no ascitic fluid available on night of admission. No
signs of heart failure noted on exam. This was ___ to med non-compliance and lack of diet restriction.
SBP negative
diuretics:
> Furosemide 40 mg PO DAILY
> Spironolactone 50 mg PO DAILY, chosen over the usual 100mg dose d/t K+ of 4.5.

CXR was wnl, UA negative, Urine culture blood culture negative.

Pt was losing excess fluid appropriately with stable lytes on the above regimen. Pt was scheduled with current PCP for ___ check upon discharge.   Pt was scheduled for new PCP with Dr. ___ at ___ and follow up in Liver clinic to schedule outpatient screening EGD and ___.

Medications on Admission:
The Preadmission Medication list is accurate and complete.
1. Furosemide 20 mg PO DAILY
2. Spironolactone 50 mg PO DAILY
3. Albuterol Inhaler 2 PUFF IH Q4H:PRN wheezing, SOB
4. Raltegravir 400 mg PO BID
5. Emtricitabine-Tenofovir (Truvada) 1 TAB PO DAILY
6. Nicotine Patch 14 mg TD DAILY
7. Ipratropium Bromide Neb 1 NEB IH Q6H SOB

Discharge Medications:
1. Albuterol Inhaler 2 PUFF IH Q4H:PRN wheezing, SOB
2. Emtricitabine-Tenofovir (Truvada) 1 TAB PO DAILY
3. Furosemide 40 mg PO DAILY
RX *furosemide 40 mg 1 tablet(s) by mouth Daily Disp #*30 Tablet
Refills:*3
4. Ipratropium Bromide Neb 1 NEB IH Q6H SOB
5. Nicotine Patch 14 mg TD DAILY
6. Raltegravir 400 mg PO BID
7. Spironolactone 50 mg PO DAILY
8. Acetaminophen 500 mg PO Q6H:PRN pain

Discharge Disposition:
Home

Discharge Diagnosis:
Ascites from Portal HTN

Discharge Condition:
Mental Status: Clear and coherent.
Level of Consciousness: Alert and interactive.
Activity Status: Ambulatory - Independent.

Discharge Instructions:
Dear Ms. ___,
It was a pleasure taking care of you! You came to us with stomach pain and worsening distension. While you were here we did a paracentesis to remove 1.5L of fluid from your belly. We also placed you on you 40 mg of Lasix and 50 mg of Aldactone to help you urinate the excess fluid still in your belly. As we discussed, everyone has a different dose of lasix required to make them urinate and it's likely that you weren't taking a high enough dose. Please take these medications daily to keep excess fluid off and eat a low salt diet. You will follow up with Dr. ___ in liver clinic and from there have your colonoscopy and EGD scheduled. Of course, we are always here if you need us.
We wish you all the best!
Your ___ Team.

Followup Instructions:
___

**LLM Input Example 2**

Name:  ___                Unit No:   ___

Admission Date:  ___            Discharge Date:   ___

Date of Birth:  ___          Sex:   F

Service: SURGERY

Allergies:
allopurinol / Statins-Hmg-Coa Reductase Inhibitors

Attending: ___.

Chief Complaint:
Left lower extremity surgical site infection

Major Surgical or Invasive Procedure:
___ Left lower extremity incision and drainage,
debridement of left foot ulcer
___ Left lower extremity washout, wound vac placement
___ Left lower extremity wound vac change, debridement
left lower extremity ulcers

History of Present Illness:
Ms. ___ Zaragosa is a ___ 60 year old female recently admitted for management of a chronic, non-healing left foot ulcer who underwent a left femoral to above knee popliteal bypass with NRGSV and left foot ulcer debridement with wound vacplacement. She was seen in clinic 3 days ago with left leg incision healing slowly and evidence of skin separation and wound infection in the left thigh. There was weeping fluid but not purulent and staples intact. There also was significant surrounding erythema and so she was sent to rehab with augmentinand was to follow-up in clinic in 2 weeks. She presents to the ED today with worsening pain and L groin to medial though wound dehiscence and purulent drainage. She is otherwise feeling well without fevers or chills, nausea or vomiting. She was admitted to the vascular surgery service for management of suspected left lower extremity surgical site infection.

Past Medical History:
PMH:
-HTN, labile
-HLD
-HYPOTHYROIDISM
-RETINAL ARTERY OCCLUSION
-Migraine

-CAD/MI (MIs in ___ and ___: This demonstrated a mid

RCA lesion which was stented with a drug-eluting stent. LAD had a proximal 30% stenosis, left circumflex had a ostial 50% stenosis. The distal RCA also had a 50% stenosis)
-CHF (EF 60-65% in ___
-OBESITY
-insulin-dependent DMII
-Gout
-Renal artery stenosis
-CKDIII
-Anemia
-afib
-Depression

PSH:
-Debridement of L foot infected ulcer
-LLE diagnostic angiogram
-L fem-AK pop bypass

Social History:
___
Family History:
Father died of colon cancer in ___.

Physical Exam:
General: NAD
CV: RRR
Pulm: No respiratory distress
Extremities: left groin wound with dressings in place. Bilateral chronic nonhealing ulcers of the lower extremities

Pertinent Results:
ADMISSION LABS:

___ 04:00PM BLOOD Neuts-86* Bands-1 Lymphs-3* Monos-9 Eos-1
Baso-0 ___ Myelos-0 AbsNeut-7.57* AbsLymp-0.26*
AbsMono-0.78 AbsEos-0.09 AbsBaso-0.00*
___ 04:00PM BLOOD ___ PTT-53.7* ___
___ 04:00PM BLOOD Glucose-118* UreaN-57* Creat-1.8* Na-139
K-4.8 Cl-97 HCO3-28 AnGap-14

DISCHARGE LABS:

___ 05:46AM BLOOD WBC-11.9* RBC-2.95* Hgb-8.7* Hct-27.5*
MCV-93 MCH-29.5 MCHC-31.6* RDW-18.4* RDWSD-59.9* Plt ___
___ 05:46AM BLOOD Plt ___
___ 05:46AM BLOOD ___ PTT-28.0 ___
___ 05:46AM BLOOD Glucose-79 UreaN-68* Creat-2.2* Na-136
K-3.7 Cl-96 HCO3-27 AnGap-13
___ 05:46AM BLOOD Calcium-7.6* Phos-5.3* Mg-2.1

Brief Hospital Course:

Ms. ___ presented on ___ to the emergency room with a concern for a surgical site infection and was given vanc/cipro/flagyl immediately. Her INR was also noted to be 5, so she received 10 of vitamin K in the emergency room. Her repeat INR was 2.3 preop. She was then taken to the operating room in the morning of ___ for a debridement and washout of the LLE. Please see OP note for more details regarding the procedure. Postoperatively, the LLE continued to exsanguinate. Cauterization and compression was done in the PACU and she was transferred to the wards. On ___ evening, she was noted to be hypotensive to SBP ___ and her Hct had drifted from 27 to 21. She was transferred to the SICU for monitoring. She received 2 units of pRBC and 1 unit of FFP along with 10 of vitamin K. Her INR was noted to be 1.7 with Hct stable at 28. Since her last echo was only done in ___, a repeat echo was done that revealed her EF to be 40%, and so she was carefully volume resuscitated in preparation for another debridement, washout and vac placement on ___. Please see op report for more details. Following her ___ postop course, her summary will be written by systems.

#NEURO: Patient was kept intubated and sedated to help facilitate multiple evaluation of her wound, however she was extubated on HD4 due to hypotension. Her pain was controlled with oxycodone and dilaudid.

#CV: Patient was noted to become transiently hypotensive to SBP ___ while she was sedated and so required levo on HD4. Her pressures improved once she was extubated. The patient remained stable from a cardiovascular standpoint; vital signs were routinely monitored.

#PULMONARY: The patient remained stable from a pulmonary standpoint; vital signs were routinely monitored. She was extubated on HD4 without issues.

#GI/GU/FEN: The patient had a foley placed intra-operatively for volume monitoring as well as to keep her incision clean. She was restarted on her home torsemide on ___. The patient was given oral diet once extubated, which she tolerated well. She was noted to have loose bowel movements and incontinent. Her C.diff was negative and so was given a flexiseal on HD3 to keep her wound clean.

#ID: The patient's fever curves were closely watched for signs of infection. She was kept on vanc/cipro/flagyl as her initial wound cultures from her initial washout was noted to be 4+GNR, 2+ GPC in pairs and chains and 1+ GPRs. On HD4, her cultures showed enterococcus and acinetobacter that were resistant to cipro and so was transitioned to ___ on HD4. ID was consulted on HD5. Given that there were no cultures showing MRSA and her vanc trough continued to be high, they recommended holding off on vanc. Her VAC was changed q3d and on her second VAC change, tissue swabs and cultures were sent that showed GPC in chains and pairs and GNRs. Updated culture data suggested VRE and daptomycin was started per ID recs. At the time of her discharge, antibiotics were discontinued according to the patient and her daughter's wishes (see below).

#HEME: Patient received several units of blood over her hospital course for low hematocrits related to bleeding from her left thigh wound. Her last transfusion was ___ and her hematocrits were stable the following two days.

#WOUNDS: The patient's left thigh wound vac was changed every ___ days. She was also found to have bilateral lower extremity pressure ulcers, more extensive on the left than the right leg. The ulcers on the left leg were found to have purulent discharge, so she was taken to the operating room again on HD9 (___) for debridement of her left lower extremity pressure ulcers. Santyl was used on these ulcers for the first 3 days post operatively. At the time of discharge to hospice, the wound vac was removed and the thigh wound was redressed with wet to dry gauze and overlying curlex.

On ___ a family meeting was held with the patient's daughter and healthcare proxy with a discussion about the lack of progression in her wounds. The following day a second meeting was held with the patient's daughter as well as representatives from palliative care, social work, case management, and vascular surgery. At that time the patient and her daughter elected to transfer the patient to a ___ facility and enact a DNR/DNI order. At the time of her discharge, the patient's vitals were stable.

Medications on Admission:
The Preadmission Medication list is accurate and complete.
1. Acetaminophen 650 mg PO Q8H
2. Albuterol Inhaler 2 PUFF IH Q4H:PRN sob
3. Aspirin 81 mg PO DAILY
4. Bisacodyl ___AILY:PRN constipation
5. BuPROPion (Sustained Release) 150 mg PO QAM
6. Digoxin 0.125 mg PO 3X/WEEK (___)
7. Docusate Sodium 100 mg PO BID constipation
8. Ezetimibe 10 mg PO DAILY
9. Febuxostat 40 mg PO DAILY
10. Ferrous Sulfate 325 mg PO BID
11. Fluticasone Propionate 110mcg 2 PUFF IH BID
12. FoLIC Acid 1 mg PO DAILY
13. HydrALAZINE 20 mg PO Q8H
14. Isosorbide Dinitrate 20 mg PO TID
15. Levothyroxine Sodium 112 mcg PO DAILY
16. Methylprednisolone 4 mg PO DAILY
17. Metoprolol Succinate XL 200 mg PO DAILY
18. Miconazole Powder 2% 1 Appl TP BID
19. nystatin 100,000 unit/gram topical ___ daily
20. Omeprazole 20 mg PO DAILY
21. OxyCODONE (Immediate Release) 10 mg PO Q4H:PRN Pain - Severe

22. Polyethylene Glycol 17 g PO DAILY constipation
23. Prasugrel 10 mg PO DAILY
24. Senna 8.6 mg PO BID:PRN constipation
25. Vitamin D ___ UNIT PO DAILY
26. Metolazone 2.5 mg PO PRN as directed by cardiologist
27. Simethicone 40-80 mg PO QID:PRN bloating
28. Spironolactone 12.5 mg PO DAILY
29. Torsemide 60 mg PO BID
30. Levofloxacin 500 mg PO Q48H foot infection

Discharge Medications:
1.  Gabapentin 100 mg PO BID
2.  Insulin SC
    Sliding Scale

Fingerstick QACHS, HS
Insulin SC Sliding Scale using REG Insulin
3.  OxyCODONE SR (OxyconTIN) 20 mg PO Q12H
RX *oxycodone 20 mg 1 tablet(s) by mouth every twelve (12) hours
Disp #*4 Tablet Refills:*0
4.  Acetaminophen 650 mg PO Q8H
5.  Albuterol Inhaler 2 PUFF IH Q4H:PRN sob
6.  Aspirin 81 mg PO DAILY
7.  Bisacodyl ___AILY:PRN constipation
8.  BuPROPion (Sustained Release) 150 mg PO QAM

9. Docusate Sodium 100 mg PO BID constipation
10. Ezetimibe 10 mg PO DAILY
11. Febuxostat 40 mg PO DAILY
12. Ferrous Sulfate 325 mg PO BID
13. Fluticasone Propionate 110mcg 2 PUFF IH BID
14. FoLIC Acid 1 mg PO DAILY
15. HydrALAZINE 20 mg PO Q8H
16. Isosorbide Dinitrate 20 mg PO TID
17. Levofloxacin 500 mg PO Q48H foot infection
18. Levothyroxine Sodium 112 mcg PO DAILY
19. Methylprednisolone 4 mg PO DAILY
20. Metoprolol Succinate XL 200 mg PO DAILY
21. Miconazole Powder 2% 1 Appl TP BID
22. nystatin 100,000 unit/gram topical ___ daily
23. Omeprazole 20 mg PO DAILY
24. OxyCODONE (Immediate Release) 10 mg PO Q4H:PRN Pain - Severe
RX *oxycodone 5 mg ___ tablet(s) by mouth every four (4) hours
Disp #*30 Tablet Refills:*0
25. Polyethylene Glycol 17 g PO DAILY constipation
26. Prasugrel 10 mg PO DAILY
27. Senna 8.6 mg PO BID:PRN constipation
28. Simethicone 40-80 mg PO QID:PRN bloating

Discharge Disposition:
Extended Care

Facility:
___

Discharge Diagnosis:
Surgical site infection of left thigh
Infection of left lower extremity pressure ulcers

Discharge Condition:
Mental Status: Confused - sometimes.
Level of Consciousness: somnolent but arousable.
Activity Status: Out of Bed with lift assistance to chair or
wheelchair.

Discharge Instructions:
Dear Ms. ___,

You were admitted to ___ on ___ with a surgical site infection of your left thigh. You were started on antibiotics and taken to the operating room for left thigh debridement and subsequently for placement of a wound vac. You were also found to have left lower extremity pressure ulcers which appeared to be infected, so you were taken back to the operating room for debridement to ensure removal of any dead or infected tissue.

At this time, you have elected to be transferred to a hospice facility. You ongoing care will be under the direction of the hospice team.

Followup Instructions:
___